\RequirePackage{tikz}
\documentclass[sn-apa,iicol]{sn-jnl}

\usepackage{amsmath,amssymb,amsfonts}
\newcommand{\norm}[1]{\left\lVert#1\right\rVert}
\usepackage{xcolor}
\usepackage{tikz}
\usetikzlibrary{matrix, calc}
\usepackage{color, colortbl}
\usepackage{caption}
\usepackage{amsmath}
\usepackage{dblfloatfix}

\jyear{2023}%
\begin{document}

\title[Article Title]{Reinforcement Learning for Shared Autonomy Drone Landings}

\author[1]{\fnm{Kal} \sur{Backman}}\email{Kal.Backman1@monash.edu}

\author[1,2]{\fnm{Dana} \sur{Kulić}}\email{Dana.Kulic@monash.edu}

\author[1]{\fnm{Hoam} \sur{Chung}}\email{Hoam.Chung@monash.edu}

\affil[1]{\orgdiv{Department of Mechanical and Aerospace Engineering}, \orgname{Monash University}, \orgaddress{\street{Wellington Road}, \city{Clayton}, \postcode{3800}, \state{Victoria}, \country{Australia}}}

\affil[2]{\orgdiv{Department of Electrical and Computer Systems Engineering}, \orgname{Monash University}, \orgaddress{\street{Wellington Road}, \city{Clayton}, \postcode{3800}, \state{Victoria}, \country{Australia}}}

\abstract{
Novice pilots find it difficult to operate and land unmanned aerial vehicles (UAVs), due to the complex UAV dynamics, challenges in depth perception, lack of expertise with the control interface and additional disturbances from the ground effect. Therefore we propose a shared autonomy approach to assist pilots in safely landing a UAV under conditions where depth perception is difficult and safe landing zones are limited. Our approach is comprised of two modules: a perception module that encodes information onto a compressed latent representation using two RGB-D cameras and a policy module that is trained with the reinforcement learning algorithm TD3 to discern the pilot’s intent and to provide control inputs that augment the user's input to safely land the UAV. 
The policy module is trained in simulation using a population of simulated users.  Simulated users are sampled from a parametric model with four parameters, which model a pilot’s tendency to conform to the assistant, proficiency, aggressiveness and speed.
We conduct a user study ($n=28$) where human participants were tasked with landing a physical UAV on one of several platforms under challenging viewing conditions. The assistant, trained with only simulated user data, improved task success rate from 51.4\% to 98.2\% despite being unaware of the human participants’ goal or the structure of the environment a priori. With the proposed assistant, regardless of prior piloting experience, participants performed with a proficiency greater than the most experienced unassisted participants.}


\maketitle

\section{Introduction}
Unmanned aerial vehicles (UAVs) are often deployed in reconnaissance, inspection and search and rescue tasks \citep{ContactInspection, SharedAutonomyPoleInspection, PVInspection, SwarmReconnaissance, SARDO}, due to their high maneuverability in full 3D space. However their mobility comes at the cost of increased teleoperation complexity. The teleoperation challenges include the pilot’s perception of the UAV’s position relative to nearby objects \citep{StereoscopicFPV, MixedRealityFPV}, coping with disturbances caused from the ground effect \citep{GroundEffect} and in understanding the mapping of teleoperation controls to UAV movement. Due to these challenges it is difficult for novice pilots to safely land a UAV, particularly in perceptually challenging environments. 

Although autonomous solutions have been proposed for UAV landing \citep{MovingVehicleLanding, MovingPlatformLanding, SDQN}, they often require prior knowledge or setup of the environment, or the specification of a known landing target. Fully autonomous solutions are unable to dynamically adapt their objective in response to context sensitive cues or external events within the environment that are not observable with the robot's sensors or a-priori specified by the designers. 
In general the difficulties associated with replicating high-level human decision making in a time and safety-critical context make teleoperation control schemes preferred over full autonomy \citep{LaserGuidedUAV, TeleOpSearch}, despite the need for expert pilots. Therefore to reduce training costs of potential pilots and improve the accessibility of UAV piloting for all, we aim to develop an assistive landing strategy that allows novice pilots to land at a proficiency at least equal to that of expert pilots.

Shared autonomy combines the control inputs of the pilot with that of an artificial intelligence to collaboratively complete a set of objectives. However, two prominent challenges when developing a shared autonomy system are predicting the intent of the user and deciding the control outputs based on the predicted user’s intent and control actions.
In our previous work \citep{Kal}, we proposed a shared autonomy system that learns to estimate the pilot's intent by training with simulated users and demonstrated it to be effective in providing assistance to human pilots landing in simulated environments. 

The contributions of this paper are the following:

\begin{itemize}
\item Rigorous validation in a physical environment with naive users.  Prior shared autonomy UAV works are either validated in simulated user studies \citep{Kal, HapticUAVInspection}, provide physical demonstrations \citep{SharedAutonomyPoleInspection, TeleOpSearch} or are evaluated with a small sample size (n=4) physical user study \citep{Reddy} where the goal is explicitly stated. The proposed work is also the first physical UAV shared autonomy system to account for multiple ambiguous goals. Prior works \citep{Reddy, SharedAutonomyPoleInspection} are limited to a single potential goal within the physical environment.

\item Reformulation of the state transition process for critics in reinforcement learning shared autonomy systems as a fully observable Markov decision process (MDP) instead of the traditional partially observable Markov decision process (POMDP) by including privileged information of the simulated user’s hidden internal state during training. The proposed reformulation significantly reduces model convergence time and improves policy performance, as MDPs offer an easier to solve problem compared to POMDPs.

\item 
Empirical comparison of the proposed approach with related shared autonomy works \cite{HindSight, Reddy}, comparing user and assistant model performance and limitations for each approach for UAV landing tasks. The experiment highlights the proposed approach’s superior performance and robustness to unseen conditions without requiring prior knowledge of the environment.

\end{itemize}

\subsection{Related work}
Previous works related to UAV landing primarily focus on full autonomy over shared autonomy. \cite{SDQN, MDPLanding} demonstrates an autonomous UAV visually seeking and landing on marked landing zones. \cite{SDQN} uses a sequence of deep Q-learning networks (DQNs) trained in an end-to-end manner in simulation. The model was transferred to reality successfully but learning the large state-space directly from images required two separate networks to be trained to counter the sample inefficiency from sparse rewards. \cite{MDPLanding} uses a Markov decision process for collision avoidance to autonomously detect landing zones and plan a path towards them. However the approach is only demonstrated in simulation and requires the landing target location to be known priori. Both approaches \citep{SDQN, MDPLanding} have access to a discrete set of actions, limiting the manoeuvrability of the UAV.  

Continuous action approaches are demonstrated in \citep{MovingPlatformDDPG, NeuralLander} where \cite{MovingPlatformDDPG} implements the continuous reinforcement learning algorithm DDPG \citep{DDPG} to land on a moving platform while \cite{NeuralLander} learns an unknown disturbance force caused by the ground effect to be fed into a non-linear trajectory tracking controller. Both approaches facilitate continuous control for UAV landing but require the desired landing position to be known a priori, limiting the feasibility in unexplored environments.

Autonomous landing zone proposal approaches have been developed to eliminate the need of explicitly specifying the landing target a priori  \citep{3DConvLandingDetection, LandingStripDetection, GaborLandingDetection}. \cite{3DConvLandingDetection} uses globally registered LiDAR point clouds to build a volumetric density map, from which a convolutional neural network is used to create a voxelised map of safe landing area probabilities. \cite{LandingStripDetection} uses publicly available population density, elevation, terrain ruggedness and land classification data in a heuristic algorithm to detect viable landing strips from which a single candidate is selected using a weighting algorithm. \cite{GaborLandingDetection} applies a series of Gabor filters to RGB images from which a Markov chain code process is used to cluster and classify pixels. However such landing proposal approaches do not take into consideration the potential objectives or preferences of the pilot and are yet to be validated for use in UAV landings.

Shared autonomy approaches circumvent the need for predefined landing zones due to the AI assistant inferring the landing position from the pilot’s actions. \cite{Reddy} implements a shared autonomy system using a model free DQN to assist participants (n=12) safely landing in the lunar lander game, where the location of safe landing zones are only known to the user. However when  tested in a user study with a physical drone (n=4), the landing pad location is included within the network’s state space.

Researchers have also proposed to estimate the intent of pilots using eye gaze tracking \citep{EyeGazeIntent} and the flight trajectory of pilots \citep{TrajectoryEstimation}. However, these models are trained in a supervised manner with ground truth labels and are yet to be implemented in the shared autonomy context.

Our previous work \citep{Kal} demonstrated a shared autonomy system where pilots were tasked to land a simulated UAV on one of several platforms with the help of an assistant.
The assistant comprised of two modules; a perception module which encodes visual information onto a latent vector and a policy module trained using the reinforcement learning algorithm TD3 \citep{TD3} to augment the pilot's actions.

Our proposed approach builds upon our prior work \citep{Kal}, making significant changes to the perception and policy modules with a focus on real-world implementation without the need for real-world data. The proposed perception module encodes information from two RGB-D cameras for greater field of view whilst maintaining a compressed latent embedding by combining the camera images with a novel camera projection model that is only attainable in simulation. For improved generality, the latent embedding encodes a segmentation map indicating potential safe landing areas in comparison to a single platform prior in \cite{Kal}. For robustness against sensor noise, noise generating functions are dynamically applied per training batch, making the module highly invariant to disturbances. 

The proposed policy module is improved with the inclusion of a LSTM-cell to address the prior inconsistency concerns shared by participants. The simulated user model is expanded to capture how a human pilot operates a joystick controller compared to keyboard presses in \cite{Kal}. Model convergence time is significantly reduced by providing hidden information of the simulated user’s intent that can only be attained in simulation to the critic, as well as a concurrent exploration training architecture. Overall task complexity is increased due to additional safe landing requirements, unrestricted platform layouts compared to one-dimensional platform configurations in \cite{Kal} and user study validation in an unseen physical environment.  We demonstrate that the proposed approach using the perceptual and policy modules trained only in simulation are able to successfully assist human pilots during physical shared autonomy drone landings.

\subsection{Problem statement and model overview}
We consider an unknown environment in which a UAV is operating, which includes a number of suitable landing locations, named \emph{landing platforms}. Our objective is to develop an assistant that aids pilots of all proficiencies to safely land a UAV on a user-selected landing platform, when neither the number nor locations of landing platforms are known to the assistant a-priori.
A landing is considered safe when the UAV remains at rest atop of a landing platform and the vertical and horizontal landing velocities are below a given threshold.
The pilot controls the UAV by providing linear XYZ velocities using a radio joystick controller for the onboard flight controller to follow. 
Due to maintaining a safe viewing distance, the pilot’s ability to perceive the depth of the UAV is hindered.

\begin{figure}
\centering
\includegraphics[width=1.0\columnwidth]{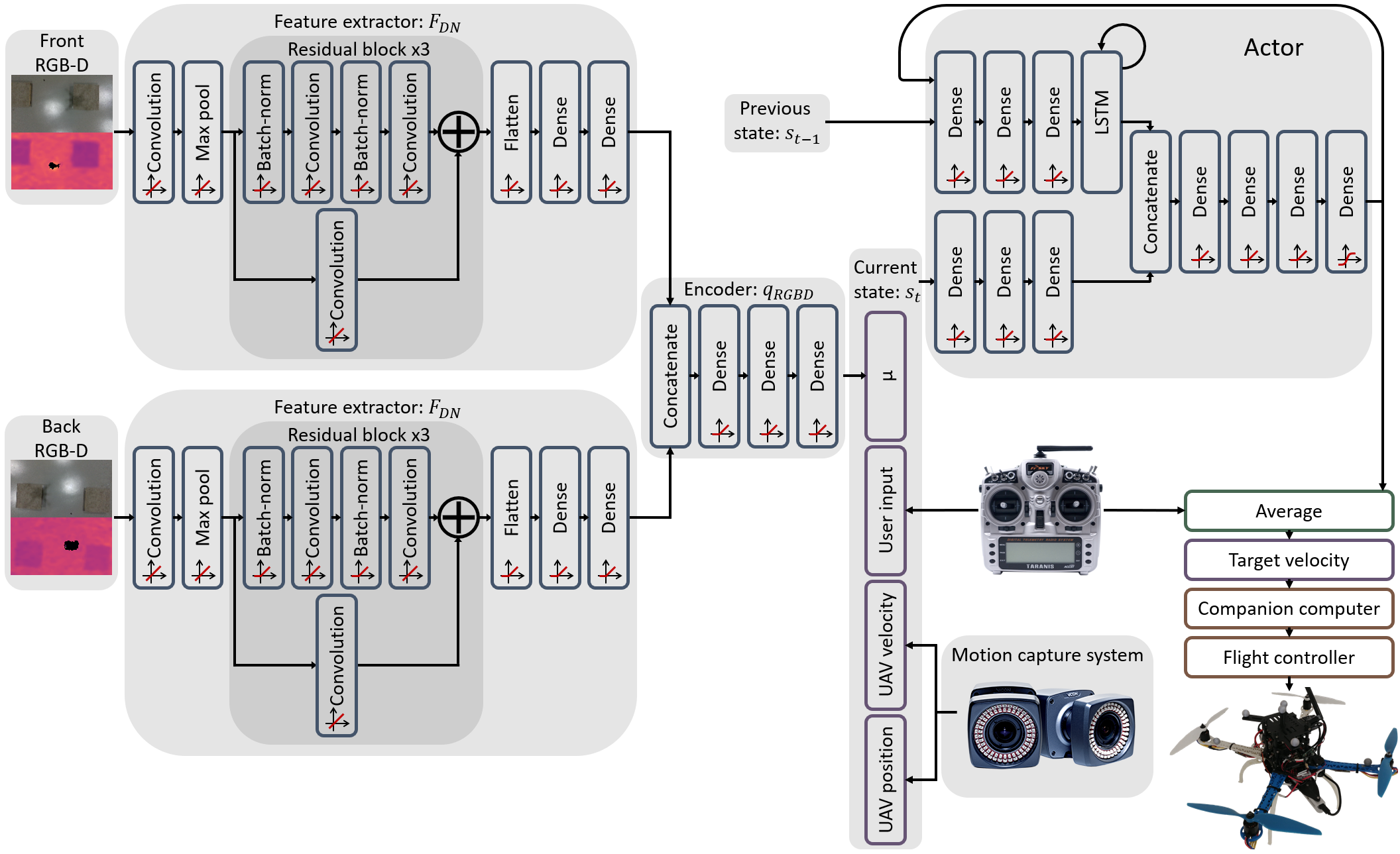}
\caption{Overview of system architecture.}
\label{ModelSummary}
\end{figure}

Our approach is summarised in Fig. \ref{ModelSummary}, and consists of  two learning components: (i) a compressed latent representation of the environment to perceive the location of potentially safe landing sites and (ii) a policy network to provide control inputs to assist the pilot in safely landing. 
The first component takes noisy images from a downwards facing stereo RGB-D camera pair and encodes information of the structure of the scene and location of safe landing zones onto a low-dimensional latent vector. 
The second component takes the current state of the UAV which includes the latent vector representation, the UAV’s dynamics and the pilot’s current action, as well as the previous state concatenated with the assistant’s previous action to output a linear XYZ target velocity. The final target velocity for the UAV’s flight controller to follow is determined by averaging the action taken by the pilot and the assistant.

\section{Learning latent space representation}

The purpose of learning a latent space embedding is to reduce the dimensionality of the input state-space the policy network learns from. As the policy network gathers experience by interacting with the environment in real time, a reduction in the dimensionality of the state space leads to accelerated convergence \citep{DimensionalityReduction}.

\cite{CM-VAE} introduces a cross-modal variational auto-encoder (CM-VAE) which trains a latent vector representation from multiple data sources. Given an input data source \(x_k\) where \(k\) represents the modality such as an RGB-D image or an arbitrary sensor reading, the encoder \(q_k\) embeds \(x_k\) onto a latent representation as a vector of means \(\mu\) and variances \(\sigma^2\) of a normal distribution. Decoder \(p_l\) samples the latent embedding \(z \sim \mathcal{N}(\mu, \sigma^2)\) then subsequently reconstructs \(z\) to the desired output \(y_l\) where \(l\) denotes the output modality. As encoder \(q_k\) maps \(x_k\) onto a normal distribution, the Kullback-Leibler divergence is used as a regularizing term to enforce that the properties of the target distribution are met, while information is encoded onto the latent vector by minimising the reconstruction loss between the predicted output \(\hat{y}_l\) and the true output \(y_l\). 

\subsection{CM-VAE implementation}
Our implementation of the CM-VAE architecture differs from our previous work \citep{Kal} by including two input RGB-D images taken from cameras situated at the front \(x_{F}\) = [\(x_{FRGB}\), \(x_{FD}\)] and back \(x_B\) = [\(x_{BRGB}\), \(x_{BD}\)] of the UAV facing downwards. The cameras are strategically placed such that the encoded field of view is maximised over the pilot’s depth axis, the axis of greatest uncertainty. The CM-VAE reconstructs two output modalities in the form of a combined depth map \(y_D\) and a binary segmentation map \(y_S\) that classifies  safe-to-land areas, where a pixel is considered safe-to-land given that all four legs of the UAV are able to contact the surface without the rest of the UAV intersecting the environment. The reconstructed output \(y_S\) replaces \(y_{XYZ}\), the relative position of the nearest landing platform in our prior work \citep{Kal} as a more general approach to defining potential landing zones, capable of representing multiple potential landing zones compared to a single platform. Both \(x_F\) and \(x_B\) are passed through a Siamese feature extraction layer \(F_{DN}\) using DroNet \citep{DroNet} from which the resultant vectors are concatenated and encoded onto the latent embedding with encoder \(q_{RGBD}\) i.e. \([\mu, \sigma ^2 ] = q_{RGBD}( [F_{DN}(x_{F}), F_{DN}(x_B)] )\). To reconstruct \(y_D\) and \(y_S\), a sample is taken from the latent embedding \(z \sim \mathcal{N}(\mu, \sigma^2)\) which is passed through decoders \(p_D\) and \(p_S\): \(\hat{y}_D = p_D(z)\) and \(\hat{y}_S = p_S(z)\).

\begin{figure}
\centering
\includegraphics[width=0.7\columnwidth]{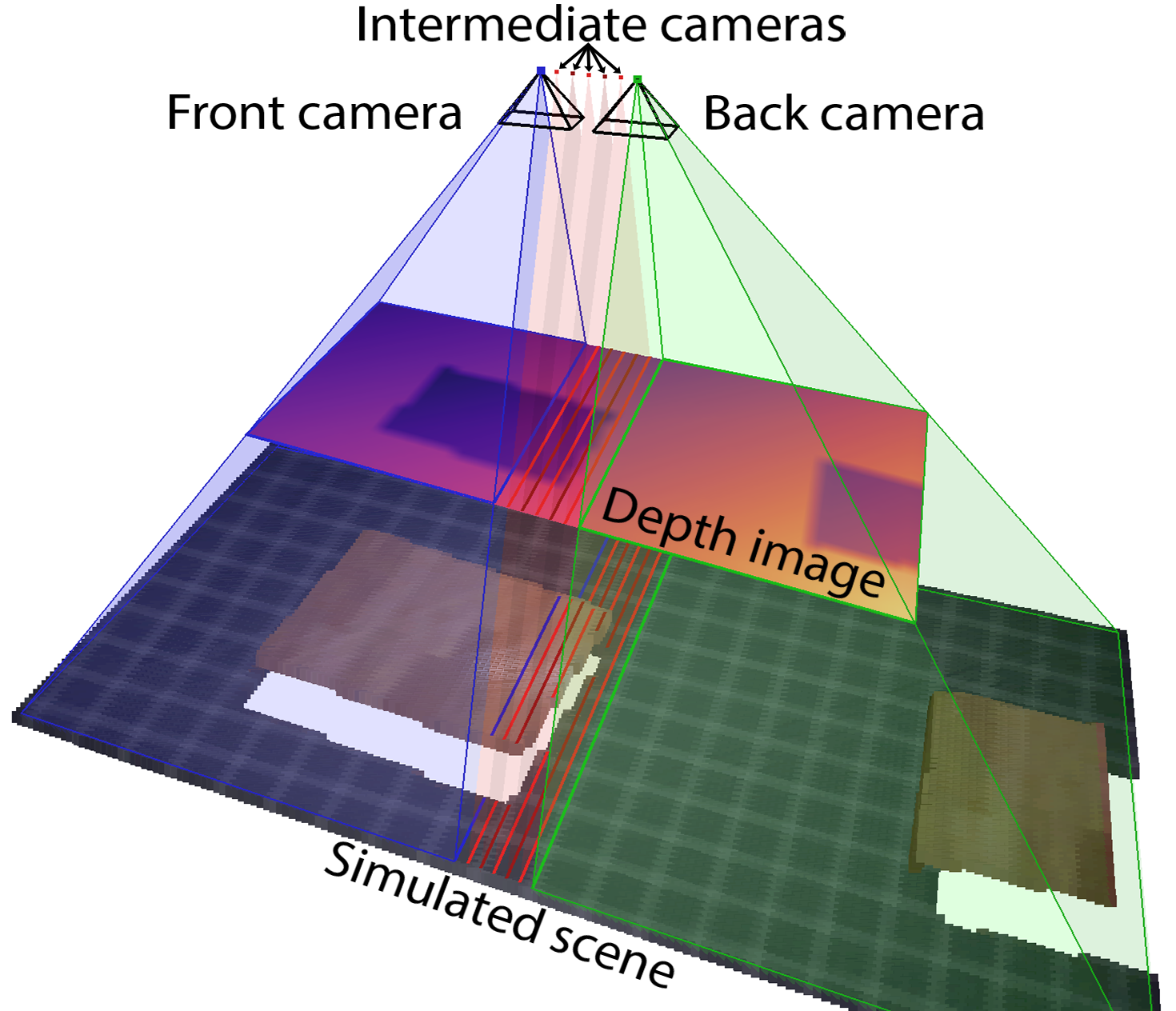}
\caption{Camera projection model used to combine stereo depth images into a single image.}
\label{DepthProjection}
\end{figure}

A consolidated latent representation for stereo RGB-D images is chosen compared to individually embedding \(x_F\) and \(x_B\) as \(\mu_F\) and \(\mu_B\), then reconstructing the depths separately as \(\hat{y}_{FD}\) and \(\hat{y}_{BD}\), in order to create a lean latent embedding by reducing redundant information that would be shared amongst both cameras' field of view. This also increases robustness against image noise as degraded visual information from one camera can be supplemented with the other. To achieve a consolidated latent representation we implement a novel camera projection model to generate \(y_D\) and \(y_S\) that encompasses the entire field of view of the input stereo image pair. The combined image is sub-divided into 3 segments, (i) front-camera and (ii) back-camera segment, which follows the standard pin-hole camera projection model split in opposite halves along the width of the image. (iii) Middle segment places \(N_c\) intermediate 1-dimensional pin-hole cameras evenly distributed between the front and back camera. The intermediate cameras share the same image height and focal lengths of the front and back cameras but have an image width of one. To generate the combined depth and segmentation map we use an image size for the front and back cameras of 141w \(\times\) 80h with \(N_c\) = 18 intermediate cameras for a combined image size of 160w \(\times\) 80h, which is subsequently down-sampled to 80w \(\times\) 40h. The combined image projection is demonstrated in Fig. \ref{DepthProjection}.

To generate data to train the network we use the AirSim \citep{AirSim} Unreal Engine plugin to capture RGB-D images of generated scenes. Each generated scene consists of \(N_p\) randomly generated landing platforms that are placed ensuring adjacent platforms are not within a minimum distance threshold. Lighting conditions are altered and textures are applied to the platforms, walls and floor from a database of 500 materials. For each scene, a total of 20 front and back RGB-D images, segmentation images and combined depth images are taken by sweeping a camera through the environment.  
As images taken from AirSim are free of noise which is not indicative of real-life sensors, a noise generating function \(G_n\) is used to generate \(x_F\) and \(x_B\) given a clean input batch of depth and RGB images. For greater domain randomisation, noised images are dynamically generated during training by parallel worker threads that send the resultant noised image batches to the network training thread.

\begin{figure}
\centering
\includegraphics[width=1.0\columnwidth]{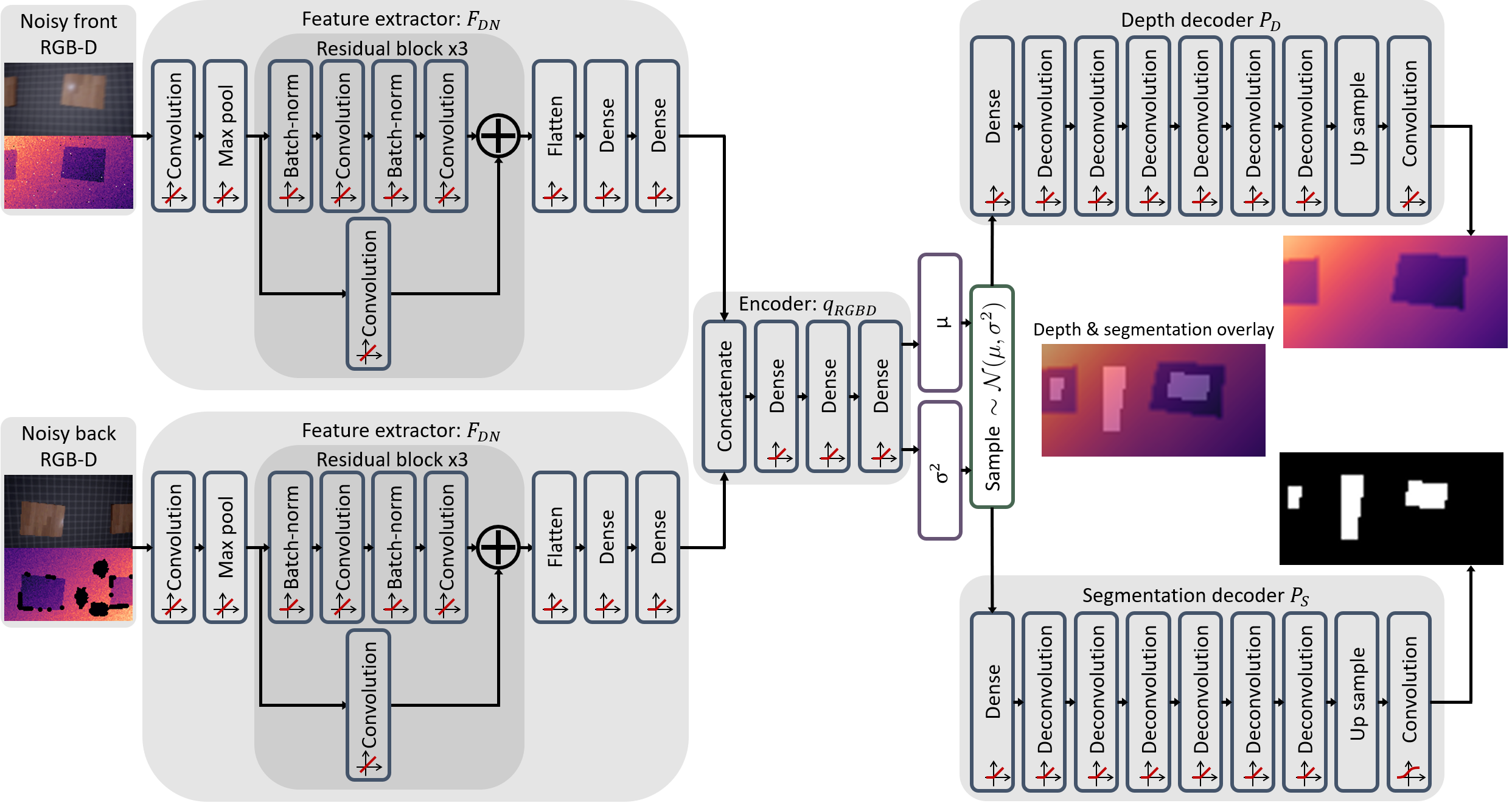}
\caption{Overview of the CM-VAE architecture.}
\label{CMVAE}
\end{figure}

Each datapoint within a training batch consists of a ground truth combined depth map, segmentation map and a pair of noisy front and back RGB-D images. Losses used can be categorized into two categories, (i) those that aim to embed information onto the latent space and (ii) those that aim to condition the latent space. For (i) we use the mean squared error between the true combined depth map and the estimated depth map \((y_D, \hat{y}_D)\) with a weighting of 1.0 and the binary cross entropy loss between the true segmentation map and the estimated segmentation map \((y_S, \hat{y}_S)\) with a weighting of 0.5. For (ii) the Kullback-Leibler divergence loss is given a weighting of 4.0.
An overview of the CM-VAE architecture can be seen in Fig. \ref{CMVAE}.

\section{Policy learning - TD3}
The aim of the policy network is to assist pilots in safely landing a UAV on a desired platform by providing a target velocity that is averaged with the pilot’s current input, which the onboard flight controller is to follow. We define our problem as a partially observable Markov decision process (POMDP) based on \cite{HindSight}’s formalization of shared autonomy systems. We define the set of all possible states as \(\mathcal{S}\), where the pilot’s goal is treated as a hidden state that must be inferred by the assistant through the observation function \(\mathcal{O} : \mathcal{S} \times \Omega \rightarrow [0, 1]\) where \(\Omega\) is the set of observations. The set of all actions available to the assistant is denoted as \(\mathcal{A}\). The state transition process is defined as a stochastic process: \(T : \mathcal{S} \times \mathcal{A} \times \mathcal{S} \rightarrow [0, 1]\), due to the uncertainty in predicting UAV dynamics from unobservable forces such as the turbulence caused from the ground effect. The reward function is defined as \(R : \mathcal{S} \times \mathcal{A} \times \mathcal{S} \rightarrow \mathbb{R}\), where the aim of reinforcement learning is to find the closest approximation to the optimal policy: \(\pi : \mathcal{S} \times \mathcal{A}\) which maximises the expected future reward with discount factor \(\gamma \in [0,1]\). 

For optimal policy approximation, we follow our previous work \citep{Kal} and use the reinforcement learning algorithm TD3 \citep{TD3} due to its model-free continuous action space control whilst alleviating the instability concerns of its predecessor DDPG \citep{DDPG}. 

As TD3 is an actor-critic approach, the actor follows the aforementioned POMDP definition where the partial observability originates from inferring the pilot’s goal without explicit definition. As the critic is an artifact of the training process that is not intended to be used during application, we model the problem for the critic as an MDP by providing the pilot’s goal information within the observation space \(\mathcal{O}\) to reduce problem complexity for the critic.

\subsection{TD3 implementation} \label{TD3Implementation}
To train the assistant we utilize AirSim as our simulation environment for modelling UAV dynamics. We generate a variety of training scenes with varying location and number of potential landing platforms. Each generated scene consists of \(N_p\) randomly generated landing platforms placed either randomly such that adjacent platforms are not within a minimum distance threshold, or are arranged in a grid-like pattern with random spacing.

\subsubsection{Simulating Users}
As training reinforcement learning algorithms takes substantial time, a population of simulated users is developed instead of training with human participants. We characterise a simulated user using four parameters: \(\alpha \in [0,1]\) describes the simulated user’s conformance to the assistant’s actions and models how likely a pilot will adopt the policy of the assistant. \(\beta \in [0,1]\) describes the user’s proficiency, defined as the ability to improve one’s estimate of the goal by perceiving the relative depth of the UAV to that of the landing platform. Both \(\alpha\) and \(\beta\) influence the simulated user’s estimate of the current goal position by:
\begin{equation}
\hat{G}_{i+1} = \hat{G}_i + \alpha \frac{a_a - a_u}{K_\alpha} + \beta \frac{G - \hat{G}_i}{K_\beta}, \label{GoalUpdateEq}
\end{equation}
where \(a_a\) \& \(a_u\) are the previous action taken by the assistant and simulated user respectively, while \(K_\alpha\) \& \(K_\beta\) are scaling constants. In our previous work \citep{Kal}, we used a 2 parameter model consisting of \(\alpha\) and \(\beta\) to model simulated users. However, these two parameters alone do not sufficiently capture the variability in user behaviour such as their desired speed and acceleration, nor model how pilots operate a physical joystick interface. Here we introduce two new parameters \(\Psi \in [0,1]\) \& \(\Phi \in [0,1]\) to model the simulated user's dexterity in implementing fine motor control of a continuous joystick controller compared to discrete keyboard inputs in \cite{Kal}. \(\Psi\) describes how aggressively a pilot accelerates their thumbs on a joystick controller, modeling their tendency to generate smooth or sharp trajectories. \(\Phi\) describes how daring a user is in terms of their maximum desired flight speeds by modeling how far a pilot is likely to push down on a joystick and their ability to provide fine adjustments when landing. 

\begin{figure}
\centering
\includegraphics[width=0.9\columnwidth]{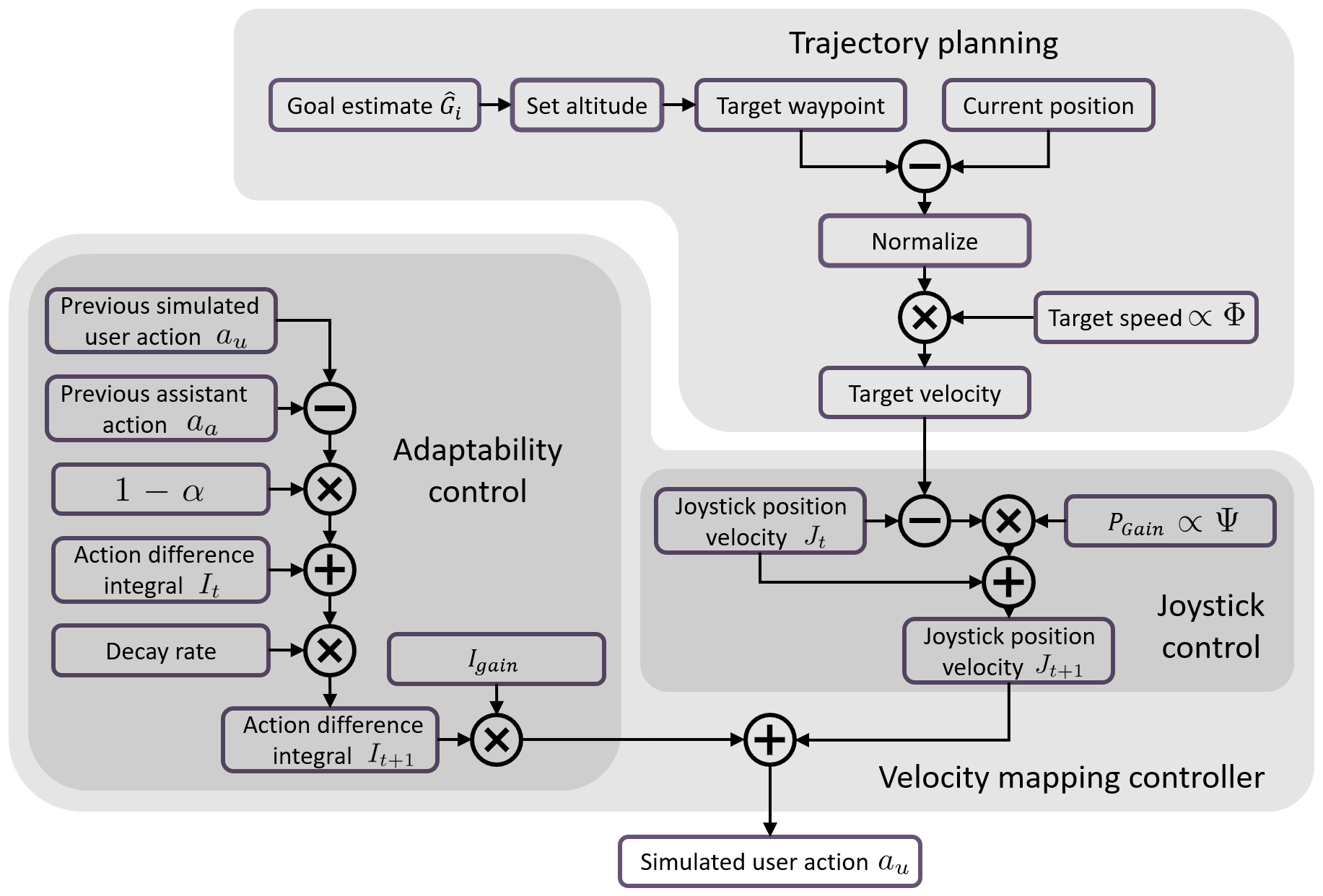}
\caption{Overview of simulated user's design}
\label{SimUser}
\end{figure}

The simulated user is modelled as a state machine comprised of two states: (i) the approach state, where the simulated user travels towards their current estimate of the goal position at a given altitude and (ii) the descent state, where the simulated user attempts to land at the current goal estimate \(\hat{G}_i\). 
The desired trajectory and velocity of the simulated user is determined by the \emph{trajectory planning} module as seen in Fig. \ref{SimUser}. The trajectory planning module is responsible for setting waypoints for the UAV to travel towards which are updated according to the current estimate of the target landing platform \(\hat{G}_i\). The desired speed at which the simulated user travels towards the set waypoint is proportional to \(\Phi\), resulting in the simulated user’s desired target velocity \(V_t\).
The output velocity \(V_t\) of the trajectory planning module is then fed into the \emph{velocity mapping controller} which determines the final output action of the simulated user as seen in Fig. \ref{SimUser}. The velocity mapping controller is split into two submodules: \emph{joystick control}, which aims to model how a human pilot’s thumbs operate a joystick controller and \emph{adaptability control}, which aims to model how a pilot may react to not traveling at their desired velocity \(V_t\) caused by disturbances from the assistant.
Given that the current joystick position corresponds to the velocity \(J_t\), the joystick control submodule will update the simulated user’s current joystick position using a proportional-controller based on the difference between its desired target velocity \(V_t\) and its current joystick input \(J_t\) i.e.
$$
J_{t+1} = J_t + (V_t – J_t)  P_{gain}
$$
where \(P_{gain} \propto \Psi \).
To adapt the simulated user’s control input in accordance with the actions taken by the assistant, the adaptability control submodule uses an integral-controller based on the previous action difference of the simulated user \(a_u\) and the assistant \(a_a\). This action difference integral \(I_t\) is updated as:
$$
I_{t+1} = I_t + (a_u – a_a)(1 - \alpha ),
$$ 
which is then subsequently decayed at each time step. The simulated user’s final output velocity command is then the sum of both the joystick control and adaptability control submodule’s outputs:
$$
a_u = J_{t+1} + I_{t+1}  I_{gain}.
$$

During training new simulated users are sampled after each landing trial where each of the four parameters: \(\alpha\), \(\beta\), \(\Psi\) and \(\Phi\) are independently sampled from a uniform distribution as well as random number generators seeded based on the current time. Each simulated user contains two random number generators to control for deterministic and non-deterministic decisions. Deterministic decisions are those that are guaranteed to occur in a specific order regardless of the actions taken by the assistant and include the initial starting altitude and whether the simulated user should fly along the principle axis or directly when approaching the target platform. Non-deterministic decisions are those that cannot be guaranteed to occur in a set order due to the actions taken by the assistant and include decisions such as pauses or altitude changes mid-flight. Two random number generators are included so that the exact characteristics and decisions made by the simulated user can be replayed when validating across multiple models for a fair comparison. Examples of the simulated user flying in various simulated environments can be seen in the supplementary video.

\subsubsection{Policy Network}
\begin{figure}[b]
\centering
\includegraphics[width=0.95\columnwidth]{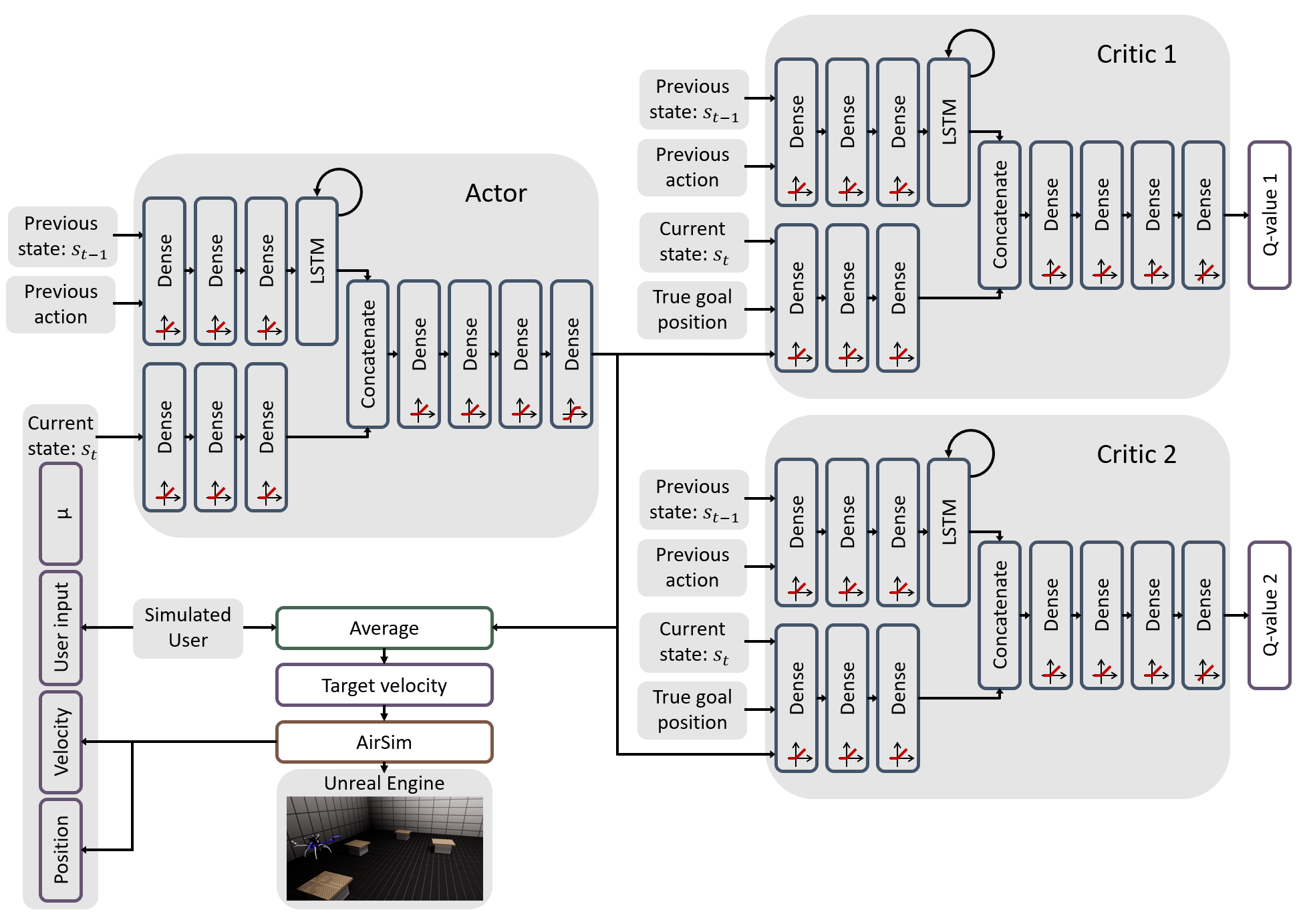}
\caption{Overview of assistant training architecture.}
\label{SimActorCritic}
\end{figure}

The actor network architecture used by the assistant to generate target velocities contains two branches of fully connected layers. The first branch extracts features of the current state \(s_t\) while the second branch extracts features of the previous state \(s_{t-1}\) which is then sequentially fed into an LSTM cell for temporal information extraction. The two branches are then concatenated and fed into additional fully connected layers to generate a vector in \(\mathbb{R}^3\),  denoting the assistant’s desired control input target velocity for the UAV. A multi-branch model is chosen compared to our prior work \citep{Kal} due to it’s poor handling of temporal information. Previously the pilot’s input was averaged over a fixed time window as a cheap means to observe the pilot’s intent over time, which occasionally led to inconsistent assistance being delivered. The actor and critic network share identical architectural designs aside from the activation layer of their final output, where the actor uses a tanh activation function scaled to generate velocity magnitudes between {\([0.0, 1.2]\) m/s} while the critic uses a linear activation function for generating the Q-value for the given state. The network architecture is summarised in Fig. \ref{SimActorCritic}.

At each iteration the actor receives three inputs: (i)  the current state vector \(s_t\) which contains the UAV position, velocity, pilot’s action and mean latent vector \(\mu\) from \(q_{RGBD}\). (ii) The previous state \(s_{t-1}\) concatenated with the action taken by the assistant in the previous iteration and (iii) the LSTM cell’s memory state for the previous iteration.

The actor is unaware of the pilot’s goal which has to be implicitly inferred through observations of the actions taken by the pilot in the context of the current state. As the critic is not required during inference, the state space for the critic can be augmented with additional information that is only attainable during training in simulation, such as the pilot’s true goal landing position. 
When training the critic we concatenate the current state \(s_t\) with the true goal position \(G\) as a means of learning with privileged information \citep{PrivilegedInformation} for decreased network convergence time.  As shown in \cite{AttentionPrivileged}, using privileged information during reinforcement learning has been demonstrated to improve learning speed and lead to impoved final performance and better generalisation.

The reward function used to train the policy includes a total of 5 terms as seen in Eq.~\ref{RewardEq}: 
\begin{equation}\label{RewardEq}
\begin{aligned}
R = k_{0}R_\mathrm{ActionDiff} + k_{1}R_\mathrm{LandingError} \\ + k_{2}R_\mathrm{SafePos} + k_{3}R_\mathrm{HVel} + k_{4}R_\mathrm{VVel}.
\end{aligned}
\end{equation} 
\(R_\mathrm{ActionDiff}\) is the difference in the action taken by the user \(a_u\) and the assistant \(a_a\): \(R_\mathrm{ActionDiff} = -\norm{a_u - a_a}\), to promote the assistant only exerting control when necessary. \(R_\mathrm{LandingError}\) is the UAV landing error towards the target platform \(G\): \(R_\mathrm{LandingError} = -\norm{G - XY}\), while \(R_\mathrm{SafePos}\) is whether the UAV landed at a safe location, defined as landing with all four legs contacting the platform (1) or not (-1). \(R_\mathrm{LandingError}\) is included to ensure the assistant lands near the platform desired by the pilot whereas \(R_\mathrm{SafePos}\) promotes safely landing on any platform. \(R_\mathrm{HVel}\) and \(R_\mathrm{VVel}\) relate to the speed of the UAV when landing in the horizontal \(v_h\) and vertical \(v_v\) direction respectively. The reward is scaled based on the height \(H\) of the UAV up to a threshold height \(H_T\) above the target platform, where the magnitude of \(R_\mathrm{HVel}\) and \(R_\mathrm{VVel}\) grows as the UAV gets closer to the platform:
\begin{equation}
R_\mathrm{HVel} = 
    \begin{cases}
        \frac{(H - H_T)}{H_T}  v_h^2  &\text{if \(H < H_{T}\)} \\
        0                                  &\text{else}
    \end{cases}
\end{equation}
 and
 \begin{equation}
 R_\mathrm{VVel} = 
    \begin{cases}
        \frac{(H - H_T)}{H_T}  v_v^2  &\text{if \(H < H_{T}\)} \\
        0                                  &\text{else}
    \end{cases}.
\end{equation}
\(R_\mathrm{HVel}\) and \(R_\mathrm{VVel}\) are included to promote a landing velocity that is safe for the UAV, where separate weightings are applied along the vertical and horizontal directions to promote the assistant in generating trajectories with minor horizontal movement when landing. The scaling term \(\frac{(H - H_T)}{H_T}\) is included to promote a smoother trajectory for the UAV compared to the assistant employing a just-in-time approach in slowing down for landing when only considering the final landing velocity in the vertical and horizontal direction. The actor aims to maximise the future discounted reward function given a discount factor of 0.99, where weighting coefficients \(k_{0}\) \textendash \ \(k_{4}\) are given values of 0.375, 12.0, 5.0, 40.0 \& 3.5 respectively.

For training, the exploration process is parallelised with 16 concurrent UAVs running across four separate instances of the Unreal Engine. Each Unreal Engine management thread collects the current state of the four UAVs within its environment, the simulated users’ actions and requests actions from the assistant from a centralised network inference thread. The returned assistant actions then have exploratory noise applied to them from an Ornstein-Uhlenbeck process which is decayed after successive epochs. The assistant actions are then averaged with the simulated users’ actions, from which normally distributed noise is then added with variance dependant on the UAV’s distance to the ground to emulate disturbances caused from the ground effect. The final target velocity is then sent to AirSim where an iteration is performed every 200ms. After each epoch the network inference thread loads the newly collected experiences from the 16 UAVs into the experience replay buffer and then performs optimisation iterations in accordance with the total exploration iterations performed in the previously completed epoch, with a batch size of 64. An epoch is considered complete when all 16 UAVs have landed or a total of 1.5 minutes have elapsed. Training and model validation results are reported in section \ref{Result_TD3}.    

\section{User study}\label{UserStudySection}
\begin{figure}[b]
\centering
\includegraphics[width=1.0\columnwidth]{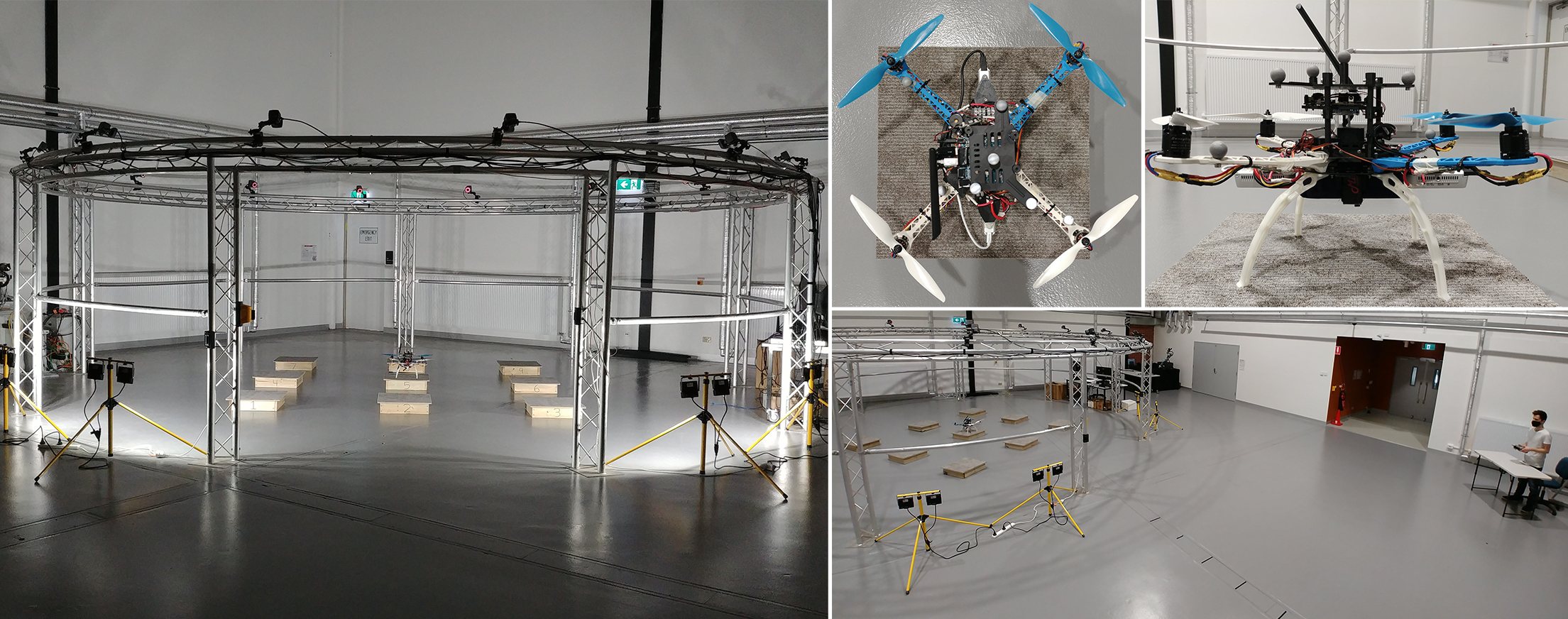}
\caption{Physical layout of user study environment. (Left) View of the arena from participant’s perspective. (Top right) UAV used in user study resting atop of landing platform. (Bottom right) Participants' relative position to the arena.}
\label{Arena}
\end{figure}

To assess the performance of the proposed assistant with human pilots, a user study was conducted. The study was approved by the Monash University Human Research Ethics Committee (MUHREC), project ID 29565. The physical arena consisted of nine labeled platforms of size 0.5\(\times\)0.5\(\times\)0.12m, arranged in a 3\(\times\)3 grid with 1.4m spacing from each platforms' centroids. Each participant performed a total of twenty landings split into a sequence of ten landings unassisted and ten landings assisted, where odd participant ID number participants completed all 10 assisted landings first while even ID number participants completed all 10 unassisted landings first. 
 During the study, participants stood as far back as possible, confined within the physical dimensions of the experiment laboratory which was measured to be 11m behind the centre platform
 and were told to remain stationary during the flight. The physical layout of the study can be seen in Fig. \ref{Arena}.

Participants flew a custom-built UAV using the Pixhawk Cube flight controller running ArduPilot. The UAV contained an Odroid-N2 as an onboard companion computer to retrieve RGB-D images from two Intel RealSense D435i cameras situated at the front and back of the UAV. The Odroid-N2 wirelessly streamed the camera feeds to a base station whilst receiving target velocities and UAV poses recorded using fourteen Bonita-10 Vicon motion capture cameras which are forwarded onto the onboard flight controller. The motion capture cameras were mounted around a 7.2m diameter circular truss, which the participants flew inside, constrained within a 6.8m diameter safe-to-fly zone by an automatic safety system which alerted and forcibly took control of the UAV if the pilot’s actions were deemed unsafe. During the study a single participant in the unassisted condition triggered the safety system due to unintentionally attempting to fly out of bounds.

Participants were initially asked to fill out a demographics survey detailing their previous experiences with UAVs and joystick controllers, after which were then shown an introductory video explaining the process of the study. Participants were then briefly given the opportunity to practice flying the UAV, where they were instructed to not perform any landings but to develop an understanding of the mapping of control inputs to the physical changes in the UAV state. After the conclusion of the practice flight, the ceiling lights were turned off and eight LED flood lights situated around the base of the truss were turned on to illuminate the arena and to prevent participants relying on shadows cast from the UAV as a substitute for depth perception. Participants were then tasked with performing the predefined sequence of ten landings either unassisted or assisted depending on their participant ID, after which they were given a NASA Task Load Index (TLX) \citep{TLX} survey detailing how they perceived the task workload for the given condition. 

To avoid the need for participants to teleoperate the UAV take-off, the first author initiated the take-off sequence and ascended the UAV to an altitude of 1.4m using a separate master RC transmitter. The participant was then instructed which platform they were to land on using the ID number inscribed on the platform. The task commenced once the participant pushed either of the joysticks. The task concluded when the UAV descended to an altitude such that the UAV’s legs would make contact with the target platform, from which the UAV’s motors were turned off and the final landing position was recorded. The sequence of ten landings were then repeated under the opposite condition of unassisted or assisted, followed by the appropriate TLX survey. Finally the participants were instructed to complete the final survey which comprised of fourteen multiple choice questions asking about specific aspects of the task, followed by seven short answer questions. The full list of survey questions can be seen in Table.~\ref{DemographicsTable}, Table.~\ref{FinalSurveyTable} \& Table.~\ref{WordedResponse}, alongside the user study results in section \ref{UserStudyResults}.

\section{Results}\label{ResultsSection}

\subsection{Perception module results}\label{PerceptionResults}
To measure the effect that domain randomization has on the CM-VAE’s ability to reconstruct real-life scenes, an experiment was conducted where three identical CM-VAE models were trained using three unique datasets. The proposed CM-VAE used to train the assistant follows the methodology outlined in section \ref{UserStudySection} and uses the complete dataset. To test the impact of noise generating functions, a model is trained using the complete dataset but images do not have noise applied to them before being fed into the network. To test the impact of training over a diverse set of environments, a third model is trained using an identical amount of unique training images used in the complete dataset, but limiting the variation of textures, lighting conditions and platform configurations. A total of 25 textures, 5\% of those used in the complete dataset, are used for the platforms and flooring. Lighting conditions are restricted to a selection of ceiling lights and platform dimensions are sampled from a discrete collection of side lengths compared to a continuous sampling in the complete dataset. Each dataset consists of a total of 80,000 examples where each model is trained over 1-million iterations.

To evaluate each model’s ability to reconstruct real-life scenes, images collected from participants in the user study are used to reconstruct the flight arena with poses collected from the motion capture system. Each model reconstructs the combined depth map given input RBD-D images taken by the front and back camera of the UAV, where the mean error is computed from the ground truth combined depth map. The ground truth combined depth map is created by using the pose of the UAV, relative camera poses to the UAVs coordinate frame and camera intrinsic parameters to project rays onto a simplified version of the environment. The simplified environment mimics the intended user study arena with platforms of size 0.5\(\times\)0.5\(\times\)0.12m arranged in a 3\(\times\)3 grid with 1.4m spacing but does not consider imperfections in the platforms or flooring. 

A total of 65338 samples were used to calculate each model's mean reconstruction error. The errors for the complete data set, limited domain randomization and no noise dataset models are 0.035m, 0.039m and 0.051m respectively compared to the average input depth map errors of 0.042m. To test if a statistically significant difference in reconstruction performance exists between the proposed model and alternative models, two-sample Welch’s t-tests are used at a 99\% confidence level. The proposed model trained over the complete dataset resulted in a statistically significant lower mean reconstruction error compared to both alternative models.

\begin{figure}
\centering
\includegraphics[width=1.0\columnwidth]{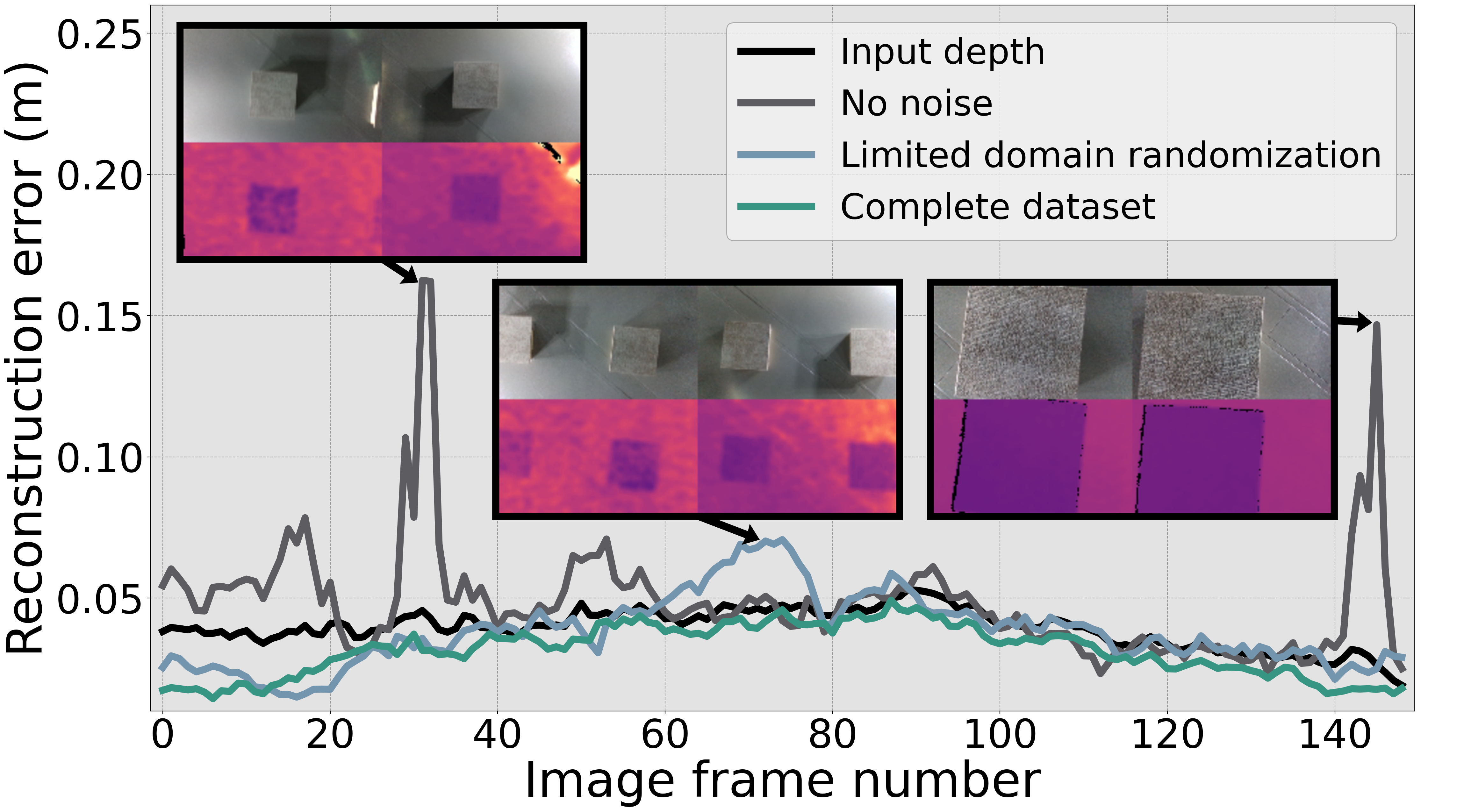}
\caption{Example CM-VAE reconstruction errors taken from a participant’s flight trajectory.
 Example input images are shown to highlight events such as spikes in the no noise model caused from input images containing missing depth values, or increases in reconstruction error for the limited domain randomization model from being unable to generalize to certain unseen configurations.  }
\label{EncoderErrors}
\end{figure}

An example plot of the reconstruction error for each input image in a participant’s trajectory can be seen in Fig. \ref{EncoderErrors}. For the model trained without noise applied to training images, large spikes in reconstruction errors can be observed in input images with missing depth values or large depth values caused from triangulating mismatched pixels. The model trained with limited domain randomization reconstructed the environment consistently however it was observed that the model had difficulty reconstructing certain sets of consecutive images. It is assumed that this is caused from difficulties associated in generalising to a broader set of unseen samples.

\begin{figure*}
\centering
\includegraphics[width=2.0\columnwidth]{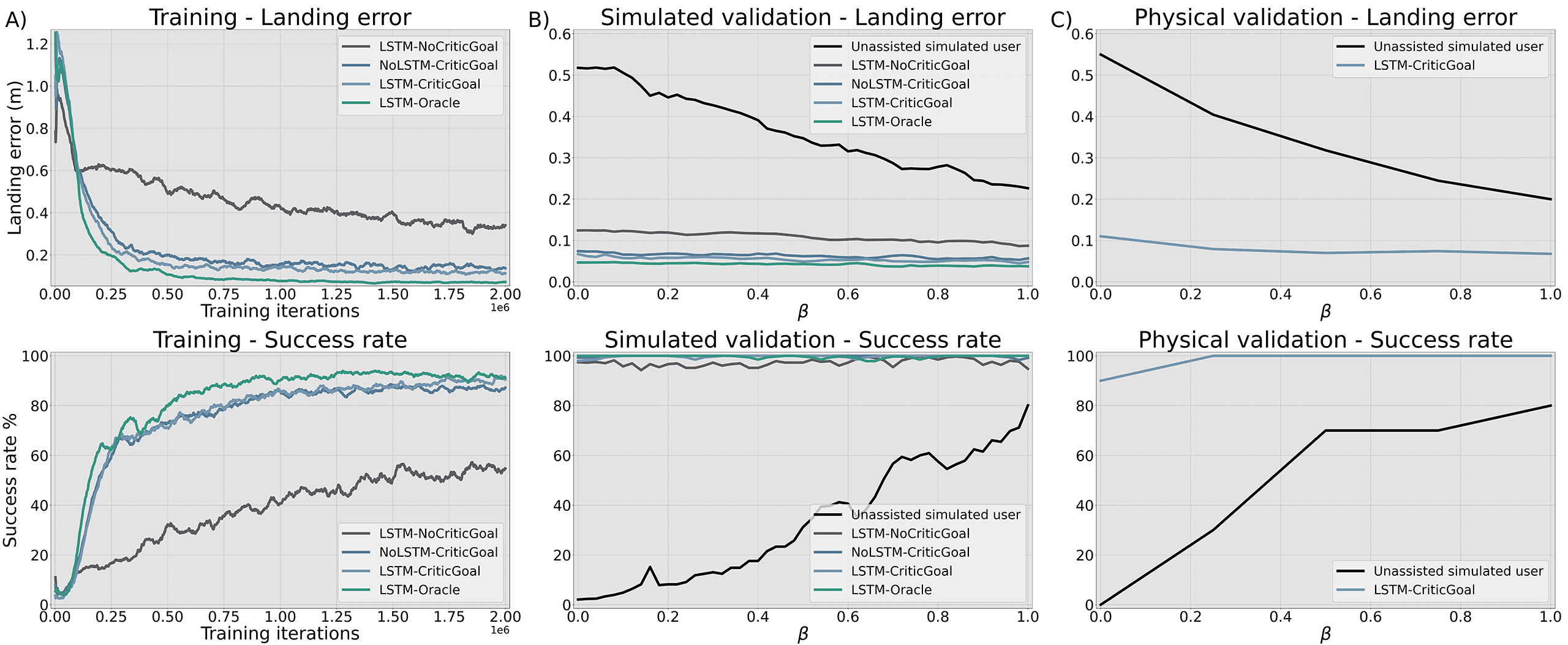}
\caption{Architecture ablation Study Results.  Each metric is averaged over the three training initialisations. For the bottom plots, a landing is considered a success given that the UAV lands on the intended landing platform with all four legs contacting the surface, whilst landing with a horizontal and vertical velocity of below 0.2 and 0.6m/s respectively.  A) Model performance whilst training; 
B) Model performance during the simulated validation task, \(\beta\) acts as a measure of the simulated user’s proficiency.
C) Model performance during the physical validation task.
}
\label{PolicyTraining}
\end{figure*}

\subsection{Policy learning results} \label{Result_TD3}
To quantify the impact of network architecture decisions, an ablation study was performed where four unique models were trained. (i) LSTM-CriticGoal, the proposed model architecture which includes an LSTM cell and where only the critic has access to the true landing position \(G\). (ii) LSTM-NoCriticGoal, where the critic is not provided with \(G\) to test what impact providing additional information to the critic has on the training process, similar to our prior work \citep{Kal}. (iii) NoLSTM-CriticGoal, where the LSTM cell is omitted from the network architecture and the fully connected layers from the current and previous state branches are directly concatenated to each other to test the impact of the LSTM cell on performance. (iv) LSTM-Oracle, where \(G\) is provided to both actor and critic to test the impact of keeping the pilot’s intent hidden from the actor. Each model was trained over three random initialisations, where each initialisation consisted of two-million training iterations. The training results can be seen in Fig. \ref{PolicyTraining}-A.

Providing the critic with the true goal information (available only in simulation) can be seen to have the greatest impact on training results. The final correct safe landing rate for the LSTM-NoCriticGoal model was 55\%, which was achievable within the first 22\% of training iterations for alternative models that included the true goal within the critic’s state space. LSTM-Oracle was the highest performing model during training, with a final average landing error 57\% lower than the average of LSTM-CriticGoal and NoLSTM-CriticGoal. Unlike the alternate models, LSTM-Oracle does not mistakenly land on incorrect platforms from wrongly inferring the simulated user’s intent.     

Each trained model was then subsequently validated on a standardised validation sequence which consisted of ten landings. The simulated environment was modeled to reflect the physical environment used in the user study, where a total of nine platforms of size 0.5\(\times\)0.5\(\times\)0.12m arranged in a 3\(\times\)3 grid with 1.4m spacing from the platform’s centroids were used. For each model, the sequence of ten landings were performed a total 51 times, where \(\beta\) was swept from 0 to 1 in 0.02 increments whilst the remaining of the simulated user’s parameters \(\alpha\), \(\Psi\) and \(\Phi\) were held constant at a value of 0.5. To ensure the simulated user’s policy remained consistent across all validated models, each of the ten landings were assigned a unique number to seed the simulated user’s random number generators. The simulated validation results can be seen in Fig. \ref{PolicyTraining}-B.

\begin{figure*}
\centering
\includegraphics[width=2.0\columnwidth]{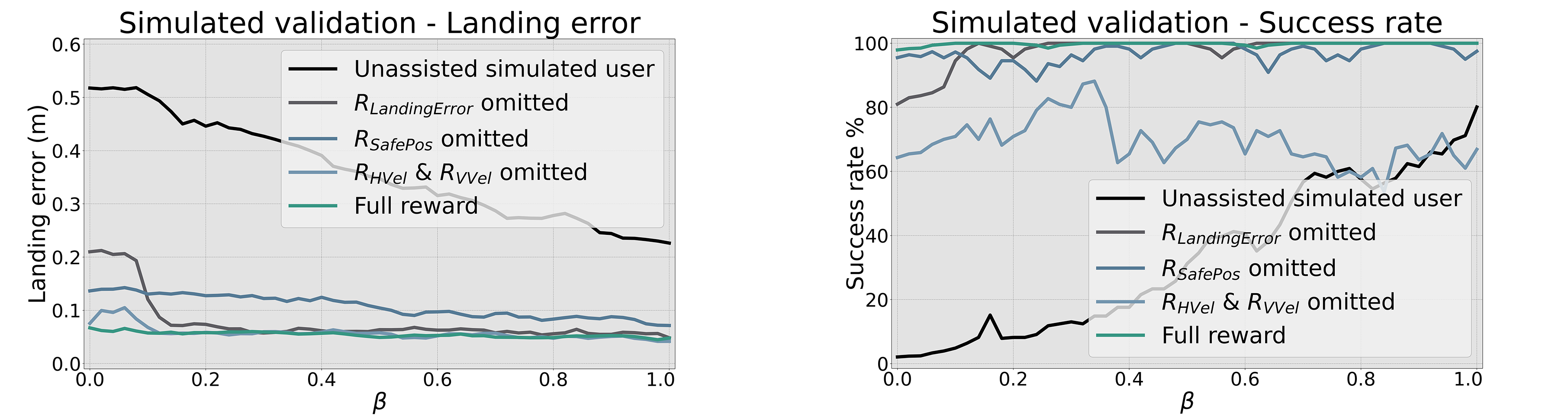}
\caption{Reward ablation study, simulated validation results }
\label{RewardCombined}
\end{figure*}

The highest performing architecture was the oracle configuration with an average landing error of 0.042m followed by LSTM-CriticGoal with 0.053m, NoLSTM-CriticGoal with 0.062m and LSTM-NoCriticGoal with 0.109m, all of which outperforming the baseline of the unassisted simulated user with an average landing error of 0.362m. Comparing the assisted approaches to that of the unassisted simulated user for both landing error and task success rate, it can be seen that the performance of the assisted approaches are invariant to that of the proficiency of the simulated user. 

To assess the performance of the proposed approach when transferring from simulation to reality, the LSTM-CriticGoal architecture was validated within the physical environment used in the user study.  The simulated user was used to pilot the UAV where the sequence of ten landings were performed for values of \(\beta\) from 0 to 1 in 0.25 increments for a total of fifty landings. \(\alpha\), \(\Psi\) and \(\Phi\) were held at constant values of 0.5 and random number generators seeded with identical values to that of the simulated validation. The results of the physical validation can be seen in Fig. \ref{PolicyTraining}-C.

The baseline performance of the unassisted simulated user in the physical validation environment was comparable to that in simulation, where the range of average landing errors with respect to \(\beta\) for the physical validation were [0.200, 0.550]m and [0.226, 0.518]m for simulated validation. Despite the proposed model LSTM-CriticGoal being trained purely on synthetic data, it achieved a landing success rate of 98\% on the physical validation sequence where only a single failed landing was recorded for a \(\beta = 0.0\) simulated user. The simulation-reality gap accounted for an additional 0.027m in the average landing error which can be attributed to a mixture of imperfect dynamic modelling of the UAV and the additional latency induced with wireless data streaming.

To assess the impact of reward function terms when training the proposed model LSTM-CriticGoal, an additional ablation study was performed where each of the terms in Eq.~\ref{RewardEq} were individually omitted, aside from \(R_\mathrm{ActionDiff}\). Each model was trained over a single random initialisation and then subjected to an identical simulated validation sequence as performed in the network architecture ablation study, the results of which can be seen in Fig.~\ref{RewardCombined}. 

Omitting \(R_\mathrm{HVel}\) and \(R_\mathrm{VVel}\) resulted in the largest decrease in success rate due to the UAV landing with unsafe velocities, however having little impact on the landing error. Omitting \(R_\mathrm{LandingError}\) did not appear to effect the performance of the assistant against simulated users of \(\beta > 0.2\), but caused the assistant to land on the incorrect platform for novice simulated users due to their larger initial errors. As \(R_\mathrm{SafePos}\) rewards the assistant for landing in any safe landing location, there is no incentive for the assistant to land at the location desired by the simulated user when \(R_\mathrm{LandingError}\) is removed. Omitting \(R_\mathrm{SafePos}\) caused the assistant to land in the general vicinity of the platform. \(R_\mathrm{SafePos}\) provides a binary reward for landing on a platform, resulting in strong gradients for the assistant to ensure a safe landing. As \(R_\mathrm{LandingError}\) is a continuous reward, a small decrease in the landing error results in a small decrease in the landing penalty. Therefore it becomes more beneficial for the assistant to focus on minimising \(R_\mathrm{HVel}\), \(R_\mathrm{VVel}\) and \(R_\mathrm{ActionDiff}\) instead of further reducing the landing error, resulting in an overall higher average landing error and decreased success in landing at a safe position.

The simulated user in our prior work \cite{Kal} used a two-parameter model which included  \(\alpha \) \& \(\beta \). To assess the impact of introducing two additional parameters \(\Psi \) \& \(\Phi \) and the velocity mapping controller, an additional ablation study was performed where the proposed model LSTM-CriticGoal was trained on simulated users using the previous two-parameter model in \cite{Kal}. To validate the difference in performance between the two assistants trained on different simulated user models, a standardized validation sequence consisting of 1,000 landings was used. The validation sequence was constructed by uniformly sampling the simulated user’s parameters \(\alpha \), \(\beta \), \(\Psi \) \& \(\Phi \) for each landing compared to the prior ablation study where \(\alpha \), \(\Psi \) \& \(\Phi \) remained constant at 0.5. Each landing in the validation sequence was performed in the simulated replica of the environment used in the physical user study, where a random start and goal platform was selected for each landing.

Both assistant models were then tested on identical validation sequences where the average success rate and landing error for the model trained using the four-parameter simulated user model was 96.4\% and 0.061m respectively. For the assistant trained on the simpler two-parameter simulated user model the average success rate and landing error was 81.6\% and 0.078m respectively.
The lower success rate from the model trained with \(\Psi \) \& \(\Phi \) omitted from its simulated user model is predominately caused from landing slightly off the platform which can be attributable to only experiencing simulated users in training that do not account for disturbances caused by the assistant from the adaptability control subsection of the velocity mapping controller module.

Accounting for \(\Psi \) \& \(\Phi \) is important in accurately reflecting human piloting characteristics as the results obtained from section \ref{UserStudyResults} show that the average non-zero action for unassisted participants ranged between 0.09m/s to 0.44m/s, demonstrating variability in the equivalent simulated user parameter \(\Phi \). While the average non-steady state acceleration in participants’ actions ranged from 0.25m/s\textsuperscript{2} to 1.17m/s\textsuperscript{2}, demonstrating the variability in the equivalent simulated user parameter \(\Psi \).

\subsection{Related work comparison}
To compare the proposed work to alternative shared autonomy works, the validation sequence used in Section \ref{Result_TD3} was applied to two comparison approaches proposed in \cite{HindSight, Reddy}. Each approach formulates a user model and assistance policy. 

\cite{HindSight} assists humans for robotic arm object grasping tasks and formulates a user model that outputs an action that directly approaches the goal location but with noise added to the output. The assistance policy requires the position of all goals within the environment to be known and computes an optimal action directed to each potential goal, where the policy’s output action is the weighted sum of each goal directed action weighted by the associated goal probability.

\cite{Reddy} assists humans in landing in the lunar lander game and formulates a simulated user model with a DQN trained using two rewards based on minimising the distance to goal and attaining landing success. To model imperfections in the optimal policy, the simulated user repeats the previous action with a fixed probability. The assistant copilot was trained using an identical DQN network following an identical reward structure.

The two comparison approaches \cite{HindSight, Reddy} were implemented following the respective author’s GitHub repositories \citep{HindSightGithub, ReddyGithub}, where modifications are made to \cite{HindSight}’s parameters to ensure appropriate scaling for the given task and environment. Implementation of \cite{Reddy}’s copilot was trained using identical inputs of the current state branch in our proposed work, while the network model was increased in depth and width to match the output branch of our proposed model. \cite{Reddy}’s DQN model required to be trained exclusively in the validation environment as training in randomly generated environments caused the model to fail to converge. The DQN was trained for two-million optimisation iterations, identical to that of our proposed work.

Each of the three assistance strategies were validated against each of the three simulated user models.  Each assistance strategy implemented their own action blending policy (combining the input of the simulated user and the assistant) as outlined within their respective works. For \cite{Reddy}, the simulated user actions were discretised into 27 actions as per our prior work \citep{Kal} for calculating the control sharing policy. Each model evaluation performed the 10 validation landings repeated a total of five times, where our proposed simulated user model had its \(\beta\) parameter swept from 0.0 to 1.0 in 0.25 increments whilst the remaining simulated users remained constant as per their implementation.

\begin{figure}[b]
\centering
\includegraphics[width=1.0\columnwidth]{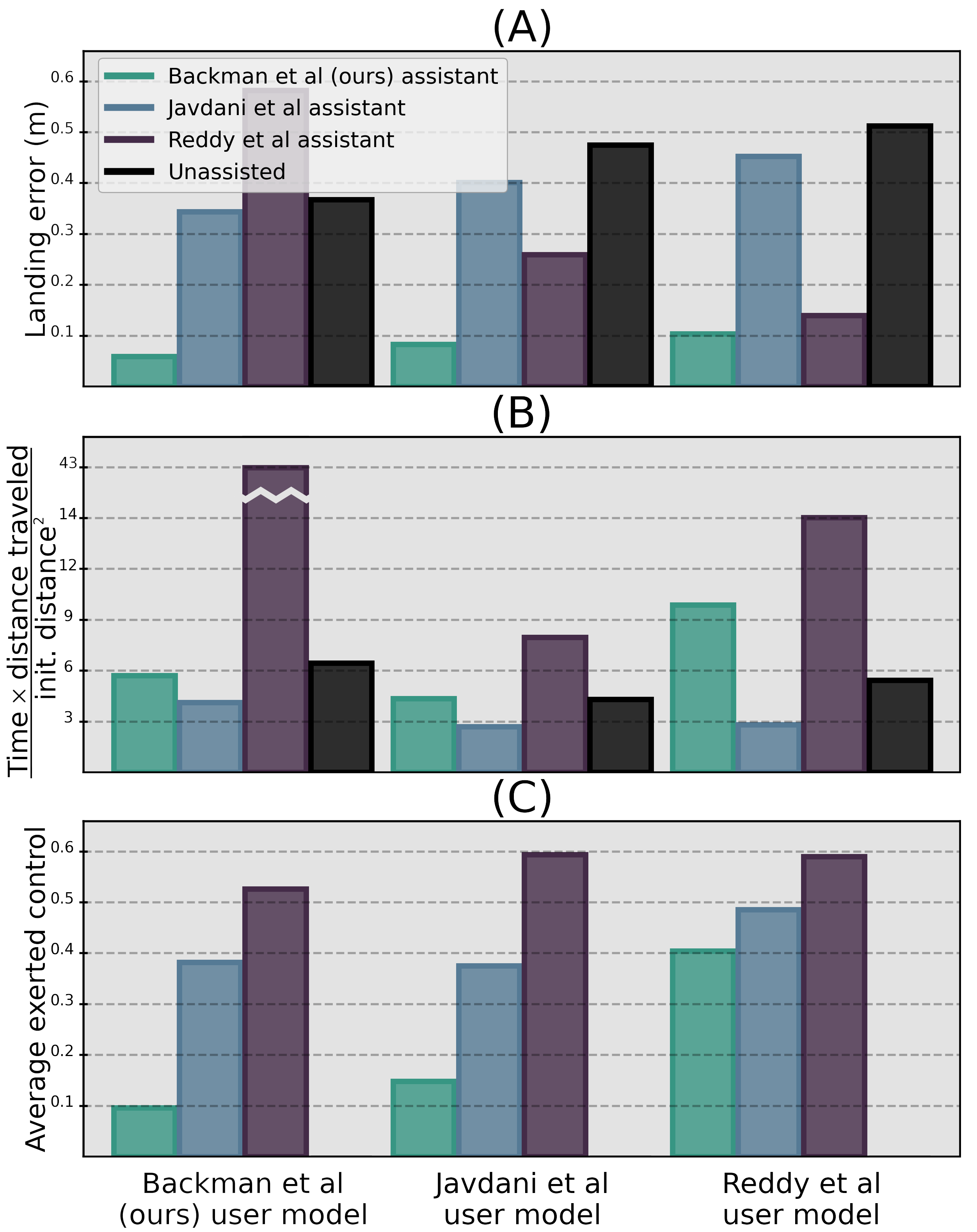}
\caption{Summary results comparing the proposed work to \cite{HindSight} and \cite{Reddy} in terms of landing error (A), trajectory efficiency (B) and assistance delivered (C). Exerted control is calculated as the Euclidean distance between the UAV’s target velocity and the simulated user’s output target velocity, which considers each assistance approach’s chosen policy blending formulation.}
\label{ComparisonImage}
\end{figure}

Summary of the simulated shared autonomy comparison results can be seen in Fig.~\ref{ComparisonImage}. Our proposed approach demonstrated the best performance in successfully landing, with an average landing error of 0.08m, compared to 0.33m and 0.40m for \cite{Reddy} and \cite{HindSight} respectively. Importantly, our proposed approach achieves the lowest landing error for all 3 user models. \cite{HindSight} generated the most efficient trajectories, defined by minimising \(\frac{\textrm{time} \times \textrm{distance travelled}}{\textrm{init. distance}^2}\), with an efficiency score of 3.20 followed by 6.64 and 22.01 by our proposed work and \cite{Reddy} respectively. In terms of the degree of control exerted onto the system, our proposed approach was shown to be the least intrusive, with an average Euclidean distance between the UAV’s target velocity and the simulated user’s output target velocity of 0.22, followed by 0.41 and 0.57 by \cite{HindSight} and \cite{Reddy} respectively. 

The main limitation of \cite{HindSight} is its lack of dynamics modelling, where the inertia of the UAV often results in overshooting of the target, which is then further exacerbated by drifting closer to incorrect goals to form greater action probabilities towards the incorrect targets. \cite{HindSight}’s success in generating efficient trajectories is due to constantly providing actions that move the UAV closer to the target, however the success comes at a cost of requiring the location of all targets within the environment as a priori, limiting its feasibility in practical settings. 

The main limitation of \cite{Reddy} is its inability to generalise to both different users or environments.  \cite{Reddy} could not be trained on multiple randomly generated environments and had poor performance when validating against our proposed simulated user model. Unlike \cite{HindSight, Reddy} user models, the proposed user model does not fly directly to the goal but performs separate approach and descent phases that more accurately model human pilot behaviour which is not modelled by the other related works, causing suboptimal performance in the unseen conditions.

The proposed model demonstrated robustness in performance across all simulated user models, despite not being trained on the other user models that perform early and aggressive descent manoeuvres. Unlike the other assistance strategies, the proposed approach does not require knowledge of the environment as a priori such as the location of all targets in \cite{HindSight} or exclusive validation arena training in \cite{Reddy}. The proposed work demonstrated the greatest performance in minimising the landing error whilst exerting the least amount of control over the system. However the lower degree of control over the system resulted in less efficient trajectories when compared to \cite{HindSight}, as the proposed approach aims to refine the pilot’s policy in order to leverage high-level human decision making, which is more effective when assisting human pilots compared to simulated users.

\subsection{User study results} \label{UserStudyResults}
28 participants completed the user study for a total of 560 landings (280 unassisted and 280 assisted) with an average study completion time of 1.5 hours per participant. A summary of key performance metrics can be viewed in Table.~\ref{SummaryResultTable}.

\definecolor{TableGray}{gray}{0.9}
\begin{table}[b]
\begin{center}
\captionof{table}{User study performance metrics summary}
\label{SummaryResultTable}
\setlength\tabcolsep{2.0pt}
\begin{tabular}{|p{4.4cm}|c|c|}
 \hline
 \multicolumn{1}{|c|}{} & \multicolumn{1}{|c}{Unassisted} & \multicolumn{1}{c|}{Assisted}\\
 \hline
  \rowcolor{TableGray} \multicolumn{1}{|p{4.4cm}|}{Success rate} & \multicolumn{1}{|c}{51.429\%} & \multicolumn{1}{c|}{\textbf{98.214\%}}\\
  \multicolumn{1}{|p{4.4cm}|}{Average landing error} & \multicolumn{1}{|c}{0.325m} & \multicolumn{1}{c|}{\textbf{0.091m}}\\
  \rowcolor{TableGray} \multicolumn{1}{|p{4.4cm}|}{Median landing error} & \multicolumn{1}{|c}{0.159m} & \multicolumn{1}{c|}{\textbf{0.081m}} \\
  \multicolumn{1}{|p{4.4cm}|}{Landing error variance} & \multicolumn{1}{|c}{0.069m$^2$} & \multicolumn{1}{c|}{\textbf{0.006m$^2$}} \\
  \rowcolor{TableGray} \multicolumn{1}{|p{4.4cm}|}{Time taken / initial starting distance} & \multicolumn{1}{|c}{10.329s/m} & \multicolumn{1}{c|}{\textbf{6.872s/m}} \\
  \multicolumn{1}{|p{4.4cm}|}{Distance traveled / initial starting distance} & \multicolumn{1}{|c}{1.705} & \multicolumn{1}{c|}{\textbf{1.522}} \\
 \hline
\end{tabular}
\end{center}
Bold values indicate the better performing condition with a statistically significant difference between the two conditions at a significance level of \(\alpha = 0.01\). 
\end{table}

The poorest performing unassisted participant scored a success rate of 10\%, whilst the highest performing unassisted participant achieved a success rate of 80\%. While in the assisted condition, the lowest success rate observed was 90\%, where the majority of assisted participants achieved a 100\% success rate. A landing was considered a success if the UAV remained at rest atop of the designated platform, landed with a horizontal and vertical velocity of below 0.2 and 0.6m/s respectively and did not engage the automatic safety system due to attempting to fly outside of the arena.

To measure the efficiency of participant’s flight strategy we assess their trajectory under two metrics, (i) the time taken to complete the task and (ii) the total distance traveled. Both metrics are normalised by the initial starting distance to account for landing sequences where the starting platform is further away from the goal platform. Whilst in the assisted condition, pilots on average required two-thirds of the time to complete the same task compared to flying unassisted. For traveled distance, assisted participants generated trajectories 18-percentage points more efficient than that of unassisted participants when comparing to the most efficient path of 1.0 for the distance traveled / initial starting distance metric.

To test if a statistically significant difference exists between the unassisted and assisted condition for the metrics provided in Table.~\ref{SummaryResultTable}, two-sample statistical tests under a 99\% confidence level are performed for each metric. To test the significance for the success rate, median landing error and landing error variance, the McNemar, Mood’s median test and Levene’s test are respectively used whilst the remaining metrics are subjected to Welch’s t-tests. All metrics were found to have a statistically significant difference between the unassisted and assisted conditions. 

To examine whether a learning effect exists from participants performing the unassisted or assisted condition first compared to the respective condition last, statistical analysis was performed on the metrics outlined in Table.~\ref{SummaryResultTable} using the previously aforementioned statistical tests under 99\% confidence. It was found for all metrics across both conditions that there was insufficient evidence to suggest that performing either unassisted or assisted condition first or last had an impact on the condition’s metrics outlined in Table.~\ref{SummaryResultTable} under 99\% confidence.

\begin{figure}[b]
\centering
\includegraphics[width=1.0\columnwidth]{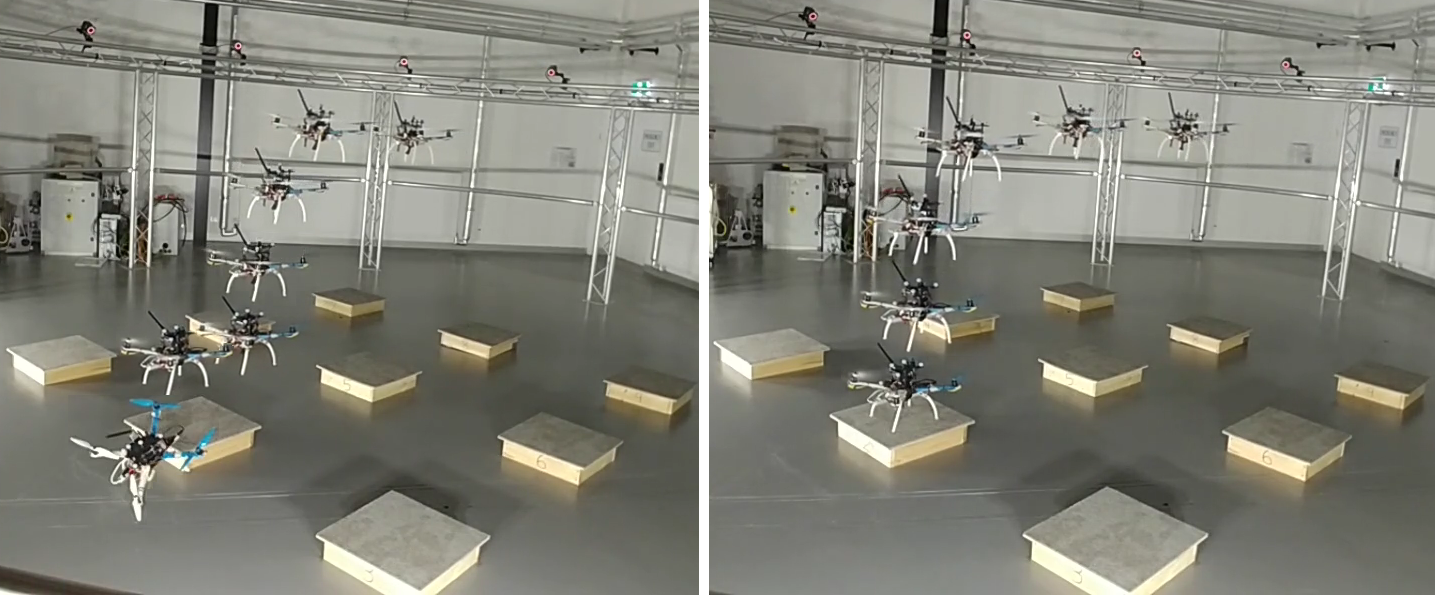}
\caption{(Left) Participant performing the task unassisted. They initially undershoot the target platform, then over-correct, failing the landing. (Right) Participant performing the task assisted. Participant instructs to prematurely descend.  The assistant alters the trajectory to land on the platform center.}
\label{TrajectoryImage}
\end{figure}

An example trajectory of a participant performing the task in the unassisted and assisted condition can be seen in Fig. \ref{TrajectoryImage}. Unassisted participants tended to undershoot their initial descent followed by showing signs of uncertainty in the UAV’s state by making multiple adjustments along the depth axis before finally landing. In the assisted condition participants continued to undershoot their initial descent however the assistant altered the trajectory to approach the platform whilst descending to ensure a safe landing. The primary cause of failures in the unassisted condition were due to the participant undershooting the landing from their initial start location (63.0\%), followed by overshooting the intended platform (32.6\%). Further unassisted and assisted landings performed by participants can be seen in the attached supplementary video.

\begin{figure}[b]
\centering
\includegraphics[width=1.0\columnwidth]{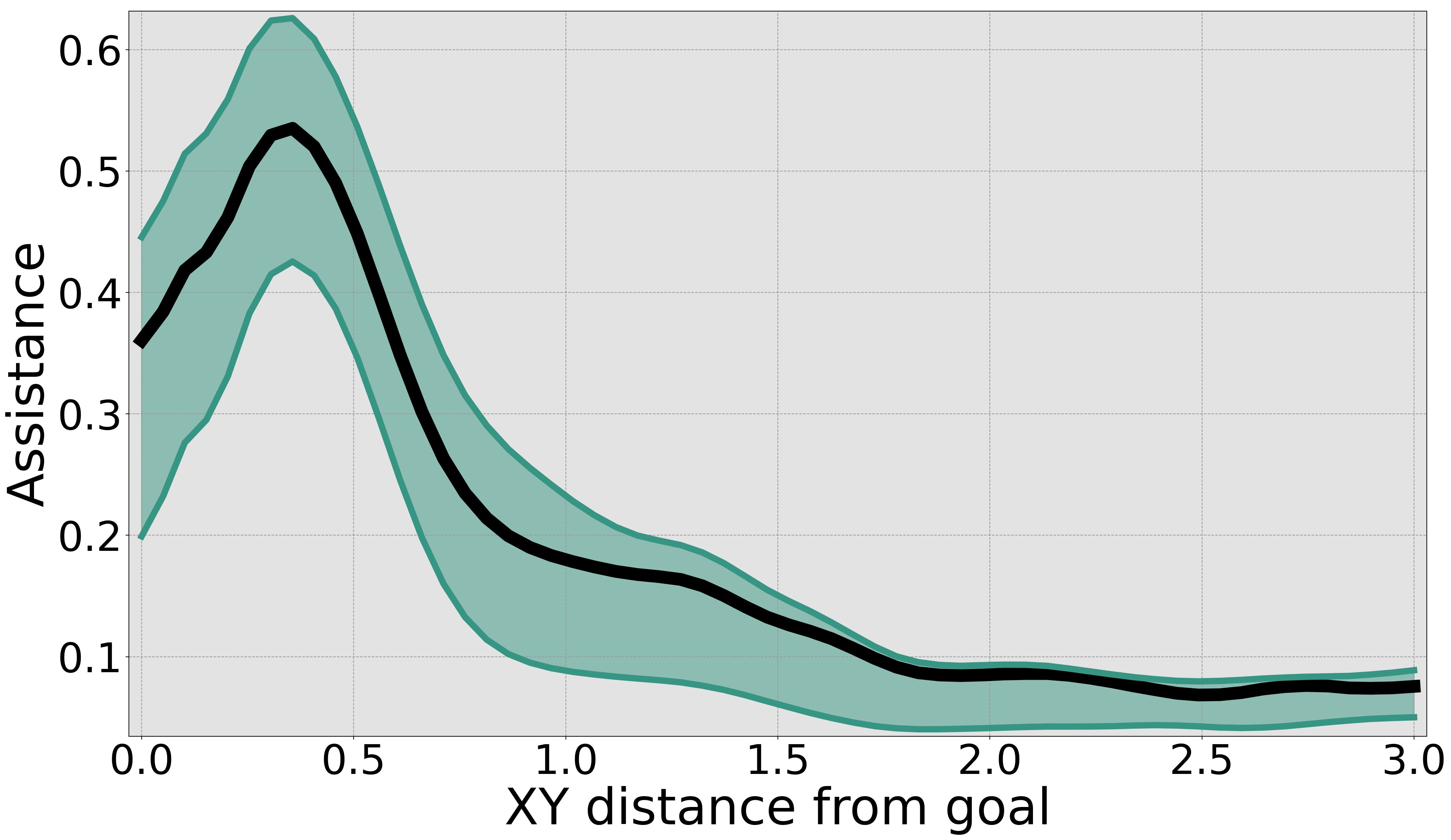}
\caption{Average assistance (black) and interquartile range (green) delivered by the assistant as a function of the XY distance from the goal location. Assistance is calculated as the XYZ distance between the pilot’s XYZ target velocity to that of the assistant’s XYZ target velocity: \(\norm{a_p - a_a}\).}
\label{AssistanceDistance}
\end{figure}

To measure how the assistance delivered by the assistant varies over trajectories, the XYZ distance between participants’ actions to that of the assistant’s action is plotted against the XY distance from the goal as seen in Fig.~\ref{AssistanceDistance}. It can be observed for distances far away from the goal (1.5m+) the assistant exerts minimal control over the UAV due to the uncertainty in the pilot’s intent. As the UAV approaches the goal location the average assistance exerted increases as the assistant’s confidence in the target increases, before declining at distances less than 0.3m due to not requiring to exert as much control to attain task success.

\begin{figure}
\centering
\includegraphics[width=1.0\columnwidth]{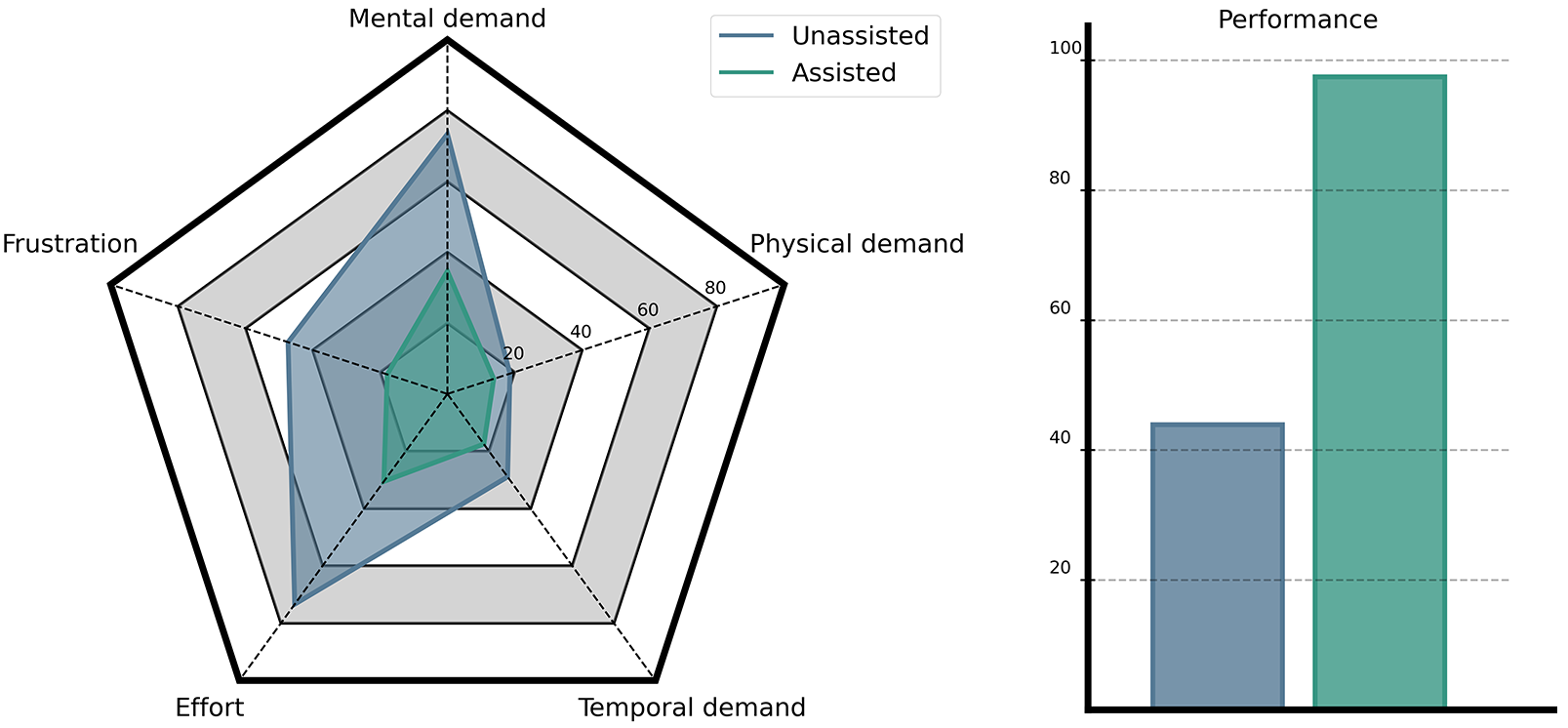}
\caption{Results of NASA TLX survey.}
\label{TLXRadar}
\end{figure}

Summary results of participants' perception of the task can be seen in Fig. \ref{TLXRadar}. From the TLX survey, participants perceived a large degree of additional effort, mental demand and frustration while performing the task in the unassisted condition and recognised a large
difference in performance whilst performing the task assisted. Participants also perceived a lower physical and temporal demand in the assisted condition albeit to a lesser extent. The TLX survey results were analysed with a Welch’s t-test under a 95\% confidence level to determine if a statistically significant difference exists between participants’ perception of the task in the unassisted and assisted condition. It was found that for all metrics aside from physical demand, a statistically significant difference exists.  

The full list of survey questions and responses can be seen in Table.~\ref{DemographicsTable}, Table.~\ref{FinalSurveyTable} and Table.~\ref{WordedResponse}. From the final survey, participants strongly agreed that with the help of the assistant, task performance and the time required to complete the task was better than that of the unassisted condition. Compared to our previous work \citep{Kal}, a stronger disparity between participant confidence was observed where participants felt less confident flying unassisted, presumably due to the presence of physical risks, whilst being more confident when flying assisted. Previously \citep{Kal} participants were neutral in terms of their trust placed onto the assistant due to inconsistencies in the degree of assistance provided as well as aggressively acting on the incorrect inference of the participants’ intent. In this work, participants' final survey results displayed a greater trust in the assistant while recognising a consistent degree of assistance and correct inference of their intent. This additional trust and consistency may explain the greater degree of overall confidence participants felt in the assisted condition compared to previously, despite the additional risks and increased task difficulty.

For the worded responses, when asked about the most difficult aspect of the task 68\% of participants made explicit references to the difficulties of estimating the depth of the UAV.  18\% of participants mentioned difficulties associated with making small adjustments especially due to ground effect as they noted that the “drone became unstable as it neared the platforms”.

For additional features that participants would like to have implemented, 21\% of participants made recommendations for a downwards facing laser light, 18\% for a downwards facing camera and 21\% for feedback via auditory or visual cues when the drone was above a platform or when the assistant was performing an action.

\newcommand{\DrawCell}[4] 
  {
  \def\TL{#1}
  \def\TR{#2}
  \def\BR{#3}
  \def\BL{#4}
  \draw (\TL) -- (\TR) -- (\BR) -- (\BL) -- (\TL);
  }
  
\newcommand{\DrawBottomlessCell}[4]
  {
  \def\TL{#1}
  \def\TR{#2}
  \def\BR{#3}
  \def\BL{#4}
  \draw (\BL)  -- (\TL) -- (\TR) -- (\BR);
  }
  
\newcommand{\DrawToplessCell}[4] 
  {
  \def\TL{#1}
  \def\TR{#2}
  \def\BR{#3}
  \def\BL{#4}
  \draw (\TL)  -- (\BL) -- (\BR) -- (\TR);
  }
  
\tikzset{ 
    DemographicSurveyTable/.style={
        matrix of nodes,
        nodes={
            rectangle,
            draw=white,
            align=center,
            text width=9.1mm
        },
        minimum height=3.0mm,
        text depth=2.5mm,
        text height=2.5mm,
        font=\footnotesize,
        nodes in empty cells,
        align=center,
        column 1/.style={
            nodes={text width=20mm, align=left}
        }
    } 
}

\begin{figure}
\captionof{table}{Participant Demographics}
\label{DemographicsTable}

\hskip-0.6em\begin{tikzpicture}
 \matrix[DemographicSurveyTable] (Demo)
 {
&&&&&\\
&&&&&\\
&&&&&\\
&&&&&\\
&&&&&\\
&&&&&\\
&&&&&\\
&&&&&\\
&&&&&\\
&&&&&\\
&&&&&\\
&&&&&\\
};
\definecolor{OddColor}{rgb}{0.3333333333333333, 0.47843137254901963, 0.5843137254901961}
\definecolor{EvenColor}{rgb}{0.21568627450980393, 0.5882352941176471, 0.5137254901960784}

\DrawBottomlessCell{Demo-1-1.north west}{Demo-1-1.north east}{Demo-1-1.south east}{Demo-1-1.south west}
\DrawToplessCell{Demo-2-1.north west}{Demo-2-1.north east}{Demo-2-1.south east}{Demo-2-1.south west}
\node[anchor = north west, align=justify, text width=20.5mm, font=\footnotesize] at (Demo-1-1.north west) {Have you any experience flying physical drones before?};
\DrawCell{Demo-1-2.north west}{Demo-1-4.north}{Demo-1-4.south}{Demo-1-2.south west}
\DrawCell{Demo-1-4.north}{Demo-1-6.north east}{Demo-1-6.south east}{Demo-1-4.south}
\DrawCell{Demo-2-2.north west}{Demo-2-4.north}{Demo-2-4.south}{Demo-2-2.south west}
\DrawCell{Demo-2-4.north}{Demo-2-6.north east}{Demo-2-6.south east}{Demo-2-4.south}
\node[anchor = north, align=justify, text width=9.1mm, font=\footnotesize] at (Demo-1-3.north) {\\ No};
\node[anchor = north, align=justify, text width=9.1mm, font=\footnotesize] at (Demo-1-5.north east) {\\ Yes};
\node[anchor = north, align=justify, text width=9.1mm, font=\footnotesize] at (Demo-2-3.north) {\\ \textbf{46\%}};
\node[anchor = north, align=justify, text width=9.1mm, font=\footnotesize] at (Demo-2-5.north east) {\\ \textbf{53\%}};
\draw[opacity=0.6,fill=OddColor] ($ (Demo-2-4.north)!1 - 46 / 100!(Demo-2-4.south) $) rectangle(Demo-2-2.south west);
\draw[opacity=0.6,fill=OddColor] ($ (Demo-2-6.north east)!1 - 53 / 100!(Demo-2-6.south east) $) rectangle(Demo-2-4.south);

\DrawBottomlessCell{Demo-3-1.north west}{Demo-3-1.north east}{Demo-3-1.south east}{Demo-3-1.south west}
\DrawToplessCell{Demo-4-1.north west}{Demo-4-1.north east}{Demo-4-1.south east}{Demo-4-1.south west}
\node[anchor = north west, align=justify, text width=20.5mm, font=\footnotesize] at (Demo-3-1.north west) {How many hours of experience have you with flying physical drones?};
\DrawCell{Demo-3-2.north west}{Demo-3-2.north east}{Demo-3-2.south east}{Demo-3-2.south west}
\DrawCell{Demo-4-2.north west}{Demo-4-2.north east}{Demo-4-2.south east}{Demo-4-2.south west}
\node[anchor = north west, align=center, text width=9.1mm, font=\tiny] at (Demo-3-2.north west) {\\ 0-5};
\node[anchor = north west, align=center, text width=9.1mm, font=\footnotesize] at (Demo-4-2.north west) {\\ \textbf{71\%}};
\draw[opacity=0.6,fill=EvenColor] ($ (Demo-4-2.north west)!1 - 71 / 100!(Demo-4-2.south west) $) rectangle(Demo-4-2.south east);
\DrawCell{Demo-3-3.north west}{Demo-3-3.north east}{Demo-3-3.south east}{Demo-3-3.south west}
\DrawCell{Demo-4-3.north west}{Demo-4-3.north east}{Demo-4-3.south east}{Demo-4-3.south west}
\node[anchor = north west, align=center, text width=9.1mm, font=\tiny] at (Demo-3-3.north west) {\\ 5-10};
\node[anchor = north west, align=center, text width=9.1mm, font=\footnotesize] at (Demo-4-3.north west) {\\ \textbf{3\%}};
\draw[opacity=0.6,fill=EvenColor] ($ (Demo-4-3.north west)!1 - 3 / 100!(Demo-4-3.south west) $) rectangle(Demo-4-3.south east);
\DrawCell{Demo-3-4.north west}{Demo-3-4.north east}{Demo-3-4.south east}{Demo-3-4.south west}
\DrawCell{Demo-4-4.north west}{Demo-4-4.north east}{Demo-4-4.south east}{Demo-4-4.south west}
\node[anchor = north west, align=center, text width=9.1mm, font=\tiny] at (Demo-3-4.north west) {\\ 10-20};
\node[anchor = north west, align=center, text width=9.1mm, font=\footnotesize] at (Demo-4-4.north west) {\\ \textbf{3\%}};
\draw[opacity=0.6,fill=EvenColor] ($ (Demo-4-4.north west)!1 - 3 / 100!(Demo-4-4.south west) $) rectangle(Demo-4-4.south east);
\DrawCell{Demo-3-5.north west}{Demo-3-5.north east}{Demo-3-5.south east}{Demo-3-5.south west}
\DrawCell{Demo-4-5.north west}{Demo-4-5.north east}{Demo-4-5.south east}{Demo-4-5.south west}
\node[anchor = north west, align=center, text width=9.1mm, font=\tiny] at (Demo-3-5.north west) {\\ 20-50};
\node[anchor = north west, align=center, text width=9.1mm, font=\footnotesize] at (Demo-4-5.north west) {\\ \textbf{10\%}};
\draw[opacity=0.6,fill=EvenColor] ($ (Demo-4-5.north west)!1 - 10 / 100!(Demo-4-5.south west) $) rectangle(Demo-4-5.south east);
\DrawCell{Demo-3-6.north west}{Demo-3-6.north east}{Demo-3-6.south east}{Demo-3-6.south west}
\DrawCell{Demo-4-6.north west}{Demo-4-6.north east}{Demo-4-6.south east}{Demo-4-6.south west}
\node[anchor = north west, align=center, text width=9.1mm, font=\tiny] at (Demo-3-6.north west) {\\ 50+};
\node[anchor = north west, align=center, text width=9.1mm, font=\footnotesize] at (Demo-4-6.north west) {\\ \textbf{10\%}};
\draw[opacity=0.6,fill=EvenColor] ($ (Demo-4-6.north west)!1 - 10 / 100!(Demo-4-6.south west) $) rectangle(Demo-4-6.south east);

\DrawBottomlessCell{Demo-5-1.north west}{Demo-5-1.north east}{Demo-5-1.south east}{Demo-5-1.south west}
\DrawToplessCell{Demo-6-1.north west}{Demo-6-1.north east}{Demo-6-1.south east}{Demo-6-1.south west}
\node[anchor = north west, align=justify, text width=20.5mm, font=\footnotesize] at (Demo-5-1.north west) {How confident are you with flying physical drones?};
\DrawCell{Demo-5-2.north west}{Demo-5-2.north east}{Demo-5-2.south east}{Demo-5-2.south west}
\DrawCell{Demo-6-2.north west}{Demo-6-2.north east}{Demo-6-2.south east}{Demo-6-2.south west}
\node[anchor = north west, align=center, text width=9.1mm, font=\tiny] at (Demo-5-2.north west) {\\ Not confident \\ 1};
\node[anchor = north west, align=center, text width=9.1mm, font=\footnotesize] at (Demo-6-2.north west) {\\ \textbf{25\%}};
\draw[opacity=0.6,fill=OddColor] ($ (Demo-6-2.north west)!1 - 25 / 100!(Demo-6-2.south west) $) rectangle(Demo-6-2.south east);
\DrawCell{Demo-5-3.north west}{Demo-5-3.north east}{Demo-5-3.south east}{Demo-5-3.south west}
\DrawCell{Demo-6-3.north west}{Demo-6-3.north east}{Demo-6-3.south east}{Demo-6-3.south west}
\node[anchor = north west, align=center, text width=9.1mm, font=\tiny] at (Demo-5-3.north west) {\\ 2};
\node[anchor = north west, align=center, text width=9.1mm, font=\footnotesize] at (Demo-6-3.north west) {\\ \textbf{35\%}};
\draw[opacity=0.6,fill=OddColor] ($ (Demo-6-3.north west)!1 - 35 / 100!(Demo-6-3.south west) $) rectangle(Demo-6-3.south east);
\DrawCell{Demo-5-4.north west}{Demo-5-4.north east}{Demo-5-4.south east}{Demo-5-4.south west}
\DrawCell{Demo-6-4.north west}{Demo-6-4.north east}{Demo-6-4.south east}{Demo-6-4.south west}
\node[anchor = north west, align=center, text width=9.1mm, font=\tiny] at (Demo-5-4.north west) {\\ 3};
\node[anchor = north west, align=center, text width=9.1mm, font=\footnotesize] at (Demo-6-4.north west) {\\ \textbf{17\%}};
\draw[opacity=0.6,fill=OddColor] ($ (Demo-6-4.north west)!1 - 17 / 100!(Demo-6-4.south west) $) rectangle(Demo-6-4.south east);
\DrawCell{Demo-5-5.north west}{Demo-5-5.north east}{Demo-5-5.south east}{Demo-5-5.south west}
\DrawCell{Demo-6-5.north west}{Demo-6-5.north east}{Demo-6-5.south east}{Demo-6-5.south west}
\node[anchor = north west, align=center, text width=9.1mm, font=\tiny] at (Demo-5-5.north west) {\\ 4};
\node[anchor = north west, align=center, text width=9.1mm, font=\footnotesize] at (Demo-6-5.north west) {\\ \textbf{14\%}};
\draw[opacity=0.6,fill=OddColor] ($ (Demo-6-5.north west)!1 - 14 / 100!(Demo-6-5.south west) $) rectangle(Demo-6-5.south east);
\DrawCell{Demo-5-6.north west}{Demo-5-6.north east}{Demo-5-6.south east}{Demo-5-6.south west}
\DrawCell{Demo-6-6.north west}{Demo-6-6.north east}{Demo-6-6.south east}{Demo-6-6.south west}
\node[anchor = north west, align=center, text width=9.1mm, font=\tiny] at (Demo-5-6.north west) {\\ Confident \\ 5};
\node[anchor = north west, align=center, text width=9.1mm, font=\footnotesize] at (Demo-6-6.north west) {\\ \textbf{7\%}};
\draw[opacity=0.6,fill=OddColor] ($ (Demo-6-6.north west)!1 - 7 / 100!(Demo-6-6.south west) $) rectangle(Demo-6-6.south east);

\DrawBottomlessCell{Demo-7-1.north west}{Demo-7-1.north east}{Demo-7-1.south east}{Demo-7-1.south west}
\DrawToplessCell{Demo-8-1.north west}{Demo-8-1.north east}{Demo-8-1.south east}{Demo-8-1.south west}
\node[anchor = north west, align=justify, text width=20.5mm, font=\footnotesize] at (Demo-7-1.north west) {How often do you play video games?};
\DrawCell{Demo-7-2.north west}{Demo-7-2.north east}{Demo-7-2.south east}{Demo-7-2.south west}
\DrawCell{Demo-8-2.north west}{Demo-8-2.north east}{Demo-8-2.south east}{Demo-8-2.south west}
\node[anchor = north west, align=center, text width=9.1mm, font=\tiny] at (Demo-7-2.north west) {\\ Never};
\node[anchor = north west, align=center, text width=9.1mm, font=\footnotesize] at (Demo-8-2.north west) {\\ \textbf{10\%}};
\draw[opacity=0.6,fill=EvenColor] ($ (Demo-8-2.north west)!1 - 10 / 100!(Demo-8-2.south west) $) rectangle(Demo-8-2.south east);
\DrawCell{Demo-7-3.north west}{Demo-7-3.north east}{Demo-7-3.south east}{Demo-7-3.south west}
\DrawCell{Demo-8-3.north west}{Demo-8-3.north east}{Demo-8-3.south east}{Demo-8-3.south west}
\node[anchor = north west, align=center, text width=9.1mm, font=\tiny] at (Demo-7-3.north west) {\\ Monthly};
\node[anchor = north west, align=center, text width=9.1mm, font=\footnotesize] at (Demo-8-3.north west) {\\ \textbf{35\%}};
\draw[opacity=0.6,fill=EvenColor] ($ (Demo-8-3.north west)!1 - 35 / 100!(Demo-8-3.south west) $) rectangle(Demo-8-3.south east);
\DrawCell{Demo-7-4.north west}{Demo-7-4.north east}{Demo-7-4.south east}{Demo-7-4.south west}
\DrawCell{Demo-8-4.north west}{Demo-8-4.north east}{Demo-8-4.south east}{Demo-8-4.south west}
\node[anchor = north west, align=center, text width=9.1mm, font=\tiny] at (Demo-7-4.north west) {\\ Weekly};
\node[anchor = north west, align=center, text width=9.1mm, font=\footnotesize] at (Demo-8-4.north west) {\\ \textbf{10\%}};
\draw[opacity=0.6,fill=EvenColor] ($ (Demo-8-4.north west)!1 - 10 / 100!(Demo-8-4.south west) $) rectangle(Demo-8-4.south east);
\DrawCell{Demo-7-5.north west}{Demo-7-5.north east}{Demo-7-5.south east}{Demo-7-5.south west}
\DrawCell{Demo-8-5.north west}{Demo-8-5.north east}{Demo-8-5.south east}{Demo-8-5.south west}
\node[anchor = north west, align=center, text width=9.1mm, font=\tiny] at (Demo-7-5.north west) {\\ Regularly};
\node[anchor = north west, align=center, text width=9.1mm, font=\footnotesize] at (Demo-8-5.north west) {\\ \textbf{21\%}};
\draw[opacity=0.6,fill=EvenColor] ($ (Demo-8-5.north west)!1 - 21 / 100!(Demo-8-5.south west) $) rectangle(Demo-8-5.south east);
\DrawCell{Demo-7-6.north west}{Demo-7-6.north east}{Demo-7-6.south east}{Demo-7-6.south west}
\DrawCell{Demo-8-6.north west}{Demo-8-6.north east}{Demo-8-6.south east}{Demo-8-6.south west}
\node[anchor = north west, align=center, text width=9.1mm, font=\tiny] at (Demo-7-6.north west) {\\ Daily};
\node[anchor = north west, align=center, text width=9.1mm, font=\footnotesize] at (Demo-8-6.north west) {\\ \textbf{21\%}};
\draw[opacity=0.6,fill=EvenColor] ($ (Demo-8-6.north west)!1 - 21 / 100!(Demo-8-6.south west) $) rectangle(Demo-8-6.south east);

\DrawBottomlessCell{Demo-9-1.north west}{Demo-9-1.north east}{Demo-9-1.south east}{Demo-9-1.south west}
\DrawToplessCell{Demo-10-1.north west}{Demo-10-1.north east}{Demo-10-1.south east}{Demo-10-1.south west}
\node[anchor = north west, align=justify, text width=20.5mm, font=\footnotesize] at (Demo-9-1.north west) {How many hours have you spent on games involving flying?};
\DrawCell{Demo-9-2.north west}{Demo-9-2.north east}{Demo-9-2.south east}{Demo-9-2.south west}
\DrawCell{Demo-10-2.north west}{Demo-10-2.north east}{Demo-10-2.south east}{Demo-10-2.south west}
\node[anchor = north west, align=center, text width=9.1mm, font=\tiny] at (Demo-9-2.north west) {\\ 0-10};
\node[anchor = north west, align=center, text width=9.1mm, font=\footnotesize] at (Demo-10-2.north west) {\\ \textbf{50\%}};
\draw[opacity=0.6,fill=OddColor] ($ (Demo-10-2.north west)!1 - 50 / 100!(Demo-10-2.south west) $) rectangle(Demo-10-2.south east);
\DrawCell{Demo-9-3.north west}{Demo-9-3.north east}{Demo-9-3.south east}{Demo-9-3.south west}
\DrawCell{Demo-10-3.north west}{Demo-10-3.north east}{Demo-10-3.south east}{Demo-10-3.south west}
\node[anchor = north west, align=center, text width=9.1mm, font=\tiny] at (Demo-9-3.north west) {\\ 10-25};
\node[anchor = north west, align=center, text width=9.1mm, font=\footnotesize] at (Demo-10-3.north west) {\\ \textbf{25\%}};
\draw[opacity=0.6,fill=OddColor] ($ (Demo-10-3.north west)!1 - 25 / 100!(Demo-10-3.south west) $) rectangle(Demo-10-3.south east);
\DrawCell{Demo-9-4.north west}{Demo-9-4.north east}{Demo-9-4.south east}{Demo-9-4.south west}
\DrawCell{Demo-10-4.north west}{Demo-10-4.north east}{Demo-10-4.south east}{Demo-10-4.south west}
\node[anchor = north west, align=center, text width=9.1mm, font=\tiny] at (Demo-9-4.north west) {\\ 25-100};
\node[anchor = north west, align=center, text width=9.1mm, font=\footnotesize] at (Demo-10-4.north west) {\\ \textbf{14\%}};
\draw[opacity=0.6,fill=OddColor] ($ (Demo-10-4.north west)!1 - 14 / 100!(Demo-10-4.south west) $) rectangle(Demo-10-4.south east);
\DrawCell{Demo-9-5.north west}{Demo-9-5.north east}{Demo-9-5.south east}{Demo-9-5.south west}
\DrawCell{Demo-10-5.north west}{Demo-10-5.north east}{Demo-10-5.south east}{Demo-10-5.south west}
\node[anchor = north west, align=center, text width=9.1mm, font=\tiny] at (Demo-9-5.north west) {\\ 100-200};
\node[anchor = north west, align=center, text width=9.1mm, font=\footnotesize] at (Demo-10-5.north west) {\\ \textbf{3\%}};
\draw[opacity=0.6,fill=OddColor] ($ (Demo-10-5.north west)!1 - 3 / 100!(Demo-10-5.south west) $) rectangle(Demo-10-5.south east);
\DrawCell{Demo-9-6.north west}{Demo-9-6.north east}{Demo-9-6.south east}{Demo-9-6.south west}
\DrawCell{Demo-10-6.north west}{Demo-10-6.north east}{Demo-10-6.south east}{Demo-10-6.south west}
\node[anchor = north west, align=center, text width=9.1mm, font=\tiny] at (Demo-9-6.north west) {\\ 200+};
\node[anchor = north west, align=center, text width=9.1mm, font=\footnotesize] at (Demo-10-6.north west) {\\ \textbf{7\%}};
\draw[opacity=0.6,fill=OddColor] ($ (Demo-10-6.north west)!1 - 7 / 100!(Demo-10-6.south west) $) rectangle(Demo-10-6.south east);

\DrawBottomlessCell{Demo-11-1.north west}{Demo-11-1.north east}{Demo-11-1.south east}{Demo-11-1.south west}
\DrawToplessCell{Demo-12-1.north west}{Demo-12-1.north east}{Demo-12-1.south east}{Demo-12-1.south west}
\node[anchor = north west, align=justify, text width=20.5mm, font=\footnotesize] at (Demo-11-1.north west) {How many hours have you spent using a joystick controller?};
\DrawCell{Demo-11-2.north west}{Demo-11-2.north east}{Demo-11-2.south east}{Demo-11-2.south west}
\DrawCell{Demo-12-2.north west}{Demo-12-2.north east}{Demo-12-2.south east}{Demo-12-2.south west}
\node[anchor = north west, align=center, text width=9.1mm, font=\tiny] at (Demo-11-2.north west) {\\ 0-10};
\node[anchor = north west, align=center, text width=9.1mm, font=\footnotesize] at (Demo-12-2.north west) {\\ \textbf{17\%}};
\draw[opacity=0.6,fill=EvenColor] ($ (Demo-12-2.north west)!1 - 17 / 100!(Demo-12-2.south west) $) rectangle(Demo-12-2.south east);
\DrawCell{Demo-11-3.north west}{Demo-11-3.north east}{Demo-11-3.south east}{Demo-11-3.south west}
\DrawCell{Demo-12-3.north west}{Demo-12-3.north east}{Demo-12-3.south east}{Demo-12-3.south west}
\node[anchor = north west, align=center, text width=9.1mm, font=\tiny] at (Demo-11-3.north west) {\\ 10-25};
\node[anchor = north west, align=center, text width=9.1mm, font=\footnotesize] at (Demo-12-3.north west) {\\ \textbf{7\%}};
\draw[opacity=0.6,fill=EvenColor] ($ (Demo-12-3.north west)!1 - 7 / 100!(Demo-12-3.south west) $) rectangle(Demo-12-3.south east);
\DrawCell{Demo-11-4.north west}{Demo-11-4.north east}{Demo-11-4.south east}{Demo-11-4.south west}
\DrawCell{Demo-12-4.north west}{Demo-12-4.north east}{Demo-12-4.south east}{Demo-12-4.south west}
\node[anchor = north west, align=center, text width=9.1mm, font=\tiny] at (Demo-11-4.north west) {\\ 25-100};
\node[anchor = north west, align=center, text width=9.1mm, font=\footnotesize] at (Demo-12-4.north west) {\\ \textbf{14\%}};
\draw[opacity=0.6,fill=EvenColor] ($ (Demo-12-4.north west)!1 - 14 / 100!(Demo-12-4.south west) $) rectangle(Demo-12-4.south east);
\DrawCell{Demo-11-5.north west}{Demo-11-5.north east}{Demo-11-5.south east}{Demo-11-5.south west}
\DrawCell{Demo-12-5.north west}{Demo-12-5.north east}{Demo-12-5.south east}{Demo-12-5.south west}
\node[anchor = north west, align=center, text width=9.1mm, font=\tiny] at (Demo-11-5.north west) {\\ 100-200};
\node[anchor = north west, align=center, text width=9.1mm, font=\footnotesize] at (Demo-12-5.north west) {\\ \textbf{32\%}};
\draw[opacity=0.6,fill=EvenColor] ($ (Demo-12-5.north west)!1 - 32 / 100!(Demo-12-5.south west) $) rectangle(Demo-12-5.south east);
\DrawCell{Demo-11-6.north west}{Demo-11-6.north east}{Demo-11-6.south east}{Demo-11-6.south west}
\DrawCell{Demo-12-6.north west}{Demo-12-6.north east}{Demo-12-6.south east}{Demo-12-6.south west}
\node[anchor = north west, align=center, text width=9.1mm, font=\tiny] at (Demo-11-6.north west) {\\ 200+};
\node[anchor = north west, align=center, text width=9.1mm, font=\footnotesize] at (Demo-12-6.north west) {\\ \textbf{28\%}};
\draw[opacity=0.6,fill=EvenColor] ($ (Demo-12-6.north west)!1 - 28 / 100!(Demo-12-6.south west) $) rectangle(Demo-12-6.south east);

\end{tikzpicture}
\end{figure}

\tikzset{ 
    FinalSurveyTable/.style={
        matrix of nodes,
        nodes={
            rectangle,
            draw=white,
            align=center,
        },
        minimum height=3.0mm,
        text depth=2.5mm,
        text height=1.5mm,
        font=\footnotesize,
        nodes in empty cells,
        align=center,
        column 1/.style={
            nodes={text width=44.5mm, align=justify}
        },
        row 1/.style={
            nodes={
                font=\bfseries, align=center, text depth=0.5mm, text height=2mm, minimum height=2mm
            }
        }
    }
}

\begin{figure}
 \captionof{table}{Final survey}
 \label{FinalSurveyTable}
\hskip-0.8em\begin{tikzpicture}

\matrix [FinalSurveyTable,text width=1.21em] (M)
{
&1&2&3&4&5\\
\textbf1: I felt confident landing unassisted&&&&&\\
\textbf2: I felt confident landing with the assistant&&&&&\\
\textbf3: I prefer landing with the help of the assistant&&&&&\\
\textbf4: I felt being unassisted gave me more freedom&&&&&\\
\textbf5: I was unsure what the assistant was trying to do&&&&&\\
\textbf6: The assistant didn’t do what I wanted it to do&&&&&\\
\textbf7: I felt like I received a lot of help from the assistant&&&&&\\
\textbf8: The assistant was consistent with the help provided&&&&&\\
\textbf9: The assistant improved the performance of the task&&&&&\\
\textbf10: The assistant made the task quicker to complete&&&&&\\
\textbf11: I trust the actions of the assistant&&&&&\\
\textbf12: The assistant and I worked well as a team&&&&&\\
\textbf13: I needed time to learn how to work with the assistant&&&&&\\
\textbf14: Given practice, I would perform better without assistance than with&&&&&\\
};

\definecolor{OddColor}{rgb}{0.3333333333333333, 0.47843137254901963, 0.5843137254901961}
\definecolor{EvenColor}{rgb}{0.21568627450980393, 0.5882352941176471, 0.5137254901960784}

\DrawCell{M-1-1.north west}{M-1-1.north east}{M-1-1.south east}{M-1-1.south west}
\DrawCell{M-1-2.north west}{M-1-2.north east}{M-1-2.south east}{M-1-2.south west}
\DrawCell{M-1-3.north west}{M-1-3.north east}{M-1-3.south east}{M-1-3.south west}
\DrawCell{M-1-4.north west}{M-1-4.north east}{M-1-4.south east}{M-1-4.south west}
\DrawCell{M-1-5.north west}{M-1-5.north east}{M-1-5.south east}{M-1-5.south west}
\DrawCell{M-1-6.north west}{M-1-6.north east}{M-1-6.south east}{M-1-6.south west}

\DrawCell{M-2-1.north west}{M-2-1.north east}{M-2-1.south east}{M-2-1.south west}
\DrawCell{M-2-2.north west}{M-2-2.north east}{M-2-2.south east}{M-2-2.south west}
\DrawCell{M-2-3.north west}{M-2-3.north east}{M-2-3.south east}{M-2-3.south west}
\DrawCell{M-2-4.north west}{M-2-4.north east}{M-2-4.south east}{M-2-4.south west}
\DrawCell{M-2-5.north west}{M-2-5.north east}{M-2-5.south east}{M-2-5.south west}
\DrawCell{M-2-6.north west}{M-2-6.north east}{M-2-6.south east}{M-2-6.south west}

\DrawCell{M-3-1.north west}{M-3-1.north east}{M-3-1.south east}{M-3-1.south west}
\DrawCell{M-3-2.north west}{M-3-2.north east}{M-3-2.south east}{M-3-2.south west}
\DrawCell{M-3-3.north west}{M-3-3.north east}{M-3-3.south east}{M-3-3.south west}
\DrawCell{M-3-4.north west}{M-3-4.north east}{M-3-4.south east}{M-3-4.south west}
\DrawCell{M-3-5.north west}{M-3-5.north east}{M-3-5.south east}{M-3-5.south west}
\DrawCell{M-3-6.north west}{M-3-6.north east}{M-3-6.south east}{M-3-6.south west}

\DrawCell{M-4-1.north west}{M-4-1.north east}{M-4-1.south east}{M-4-1.south west}
\DrawCell{M-4-2.north west}{M-4-2.north east}{M-4-2.south east}{M-4-2.south west}
\DrawCell{M-4-3.north west}{M-4-3.north east}{M-4-3.south east}{M-4-3.south west}
\DrawCell{M-4-4.north west}{M-4-4.north east}{M-4-4.south east}{M-4-4.south west}
\DrawCell{M-4-5.north west}{M-4-5.north east}{M-4-5.south east}{M-4-5.south west}
\DrawCell{M-4-6.north west}{M-4-6.north east}{M-4-6.south east}{M-4-6.south west}

\DrawCell{M-5-1.north west}{M-5-1.north east}{M-5-1.south east}{M-5-1.south west}
\DrawCell{M-5-2.north west}{M-5-2.north east}{M-5-2.south east}{M-5-2.south west}
\DrawCell{M-5-3.north west}{M-5-3.north east}{M-5-3.south east}{M-5-3.south west}
\DrawCell{M-5-4.north west}{M-5-4.north east}{M-5-4.south east}{M-5-4.south west}
\DrawCell{M-5-5.north west}{M-5-5.north east}{M-5-5.south east}{M-5-5.south west}
\DrawCell{M-5-6.north west}{M-5-6.north east}{M-5-6.south east}{M-5-6.south west}

\DrawCell{M-6-1.north west}{M-6-1.north east}{M-6-1.south east}{M-6-1.south west}
\DrawCell{M-6-2.north west}{M-6-2.north east}{M-6-2.south east}{M-6-2.south west}
\DrawCell{M-6-3.north west}{M-6-3.north east}{M-6-3.south east}{M-6-3.south west}
\DrawCell{M-6-4.north west}{M-6-4.north east}{M-6-4.south east}{M-6-4.south west}
\DrawCell{M-6-5.north west}{M-6-5.north east}{M-6-5.south east}{M-6-5.south west}
\DrawCell{M-6-6.north west}{M-6-6.north east}{M-6-6.south east}{M-6-6.south west}

\DrawCell{M-7-1.north west}{M-7-1.north east}{M-7-1.south east}{M-7-1.south west}
\DrawCell{M-7-2.north west}{M-7-2.north east}{M-7-2.south east}{M-7-2.south west}
\DrawCell{M-7-3.north west}{M-7-3.north east}{M-7-3.south east}{M-7-3.south west}
\DrawCell{M-7-4.north west}{M-7-4.north east}{M-7-4.south east}{M-7-4.south west}
\DrawCell{M-7-5.north west}{M-7-5.north east}{M-7-5.south east}{M-7-5.south west}
\DrawCell{M-7-6.north west}{M-7-6.north east}{M-7-6.south east}{M-7-6.south west}

\DrawCell{M-8-1.north west}{M-8-1.north east}{M-8-1.south east}{M-8-1.south west}
\DrawCell{M-8-2.north west}{M-8-2.north east}{M-8-2.south east}{M-8-2.south west}
\DrawCell{M-8-3.north west}{M-8-3.north east}{M-8-3.south east}{M-8-3.south west}
\DrawCell{M-8-4.north west}{M-8-4.north east}{M-8-4.south east}{M-8-4.south west}
\DrawCell{M-8-5.north west}{M-8-5.north east}{M-8-5.south east}{M-8-5.south west}
\DrawCell{M-8-6.north west}{M-8-6.north east}{M-8-6.south east}{M-8-6.south west}

\DrawCell{M-9-1.north west}{M-9-1.north east}{M-9-1.south east}{M-9-1.south west}
\DrawCell{M-9-2.north west}{M-9-2.north east}{M-9-2.south east}{M-9-2.south west}
\DrawCell{M-9-3.north west}{M-9-3.north east}{M-9-3.south east}{M-9-3.south west}
\DrawCell{M-9-4.north west}{M-9-4.north east}{M-9-4.south east}{M-9-4.south west}
\DrawCell{M-9-5.north west}{M-9-5.north east}{M-9-5.south east}{M-9-5.south west}
\DrawCell{M-9-6.north west}{M-9-6.north east}{M-9-6.south east}{M-9-6.south west}

\DrawCell{M-10-1.north west}{M-10-1.north east}{M-10-1.south east}{M-10-1.south west}
\DrawCell{M-10-2.north west}{M-10-2.north east}{M-10-2.south east}{M-10-2.south west}
\DrawCell{M-10-3.north west}{M-10-3.north east}{M-10-3.south east}{M-10-3.south west}
\DrawCell{M-10-4.north west}{M-10-4.north east}{M-10-4.south east}{M-10-4.south west}
\DrawCell{M-10-5.north west}{M-10-5.north east}{M-10-5.south east}{M-10-5.south west}
\DrawCell{M-10-6.north west}{M-10-6.north east}{M-10-6.south east}{M-10-6.south west}

\DrawCell{M-11-1.north west}{M-11-1.north east}{M-11-1.south east}{M-11-1.south west}
\DrawCell{M-11-2.north west}{M-11-2.north east}{M-11-2.south east}{M-11-2.south west}
\DrawCell{M-11-3.north west}{M-11-3.north east}{M-11-3.south east}{M-11-3.south west}
\DrawCell{M-11-4.north west}{M-11-4.north east}{M-11-4.south east}{M-11-4.south west}
\DrawCell{M-11-5.north west}{M-11-5.north east}{M-11-5.south east}{M-11-5.south west}
\DrawCell{M-11-6.north west}{M-11-6.north east}{M-11-6.south east}{M-11-6.south west}

\DrawCell{M-12-1.north west}{M-12-1.north east}{M-12-1.south east}{M-12-1.south west}
\DrawCell{M-12-2.north west}{M-12-2.north east}{M-12-2.south east}{M-12-2.south west}
\DrawCell{M-12-3.north west}{M-12-3.north east}{M-12-3.south east}{M-12-3.south west}
\DrawCell{M-12-4.north west}{M-12-4.north east}{M-12-4.south east}{M-12-4.south west}
\DrawCell{M-12-5.north west}{M-12-5.north east}{M-12-5.south east}{M-12-5.south west}
\DrawCell{M-12-6.north west}{M-12-6.north east}{M-12-6.south east}{M-12-6.south west}

\DrawCell{M-13-1.north west}{M-13-1.north east}{M-13-1.south east}{M-13-1.south west}
\DrawCell{M-13-2.north west}{M-13-2.north east}{M-13-2.south east}{M-13-2.south west}
\DrawCell{M-13-3.north west}{M-13-3.north east}{M-13-3.south east}{M-13-3.south west}
\DrawCell{M-13-4.north west}{M-13-4.north east}{M-13-4.south east}{M-13-4.south west}
\DrawCell{M-13-5.north west}{M-13-5.north east}{M-13-5.south east}{M-13-5.south west}
\DrawCell{M-13-6.north west}{M-13-6.north east}{M-13-6.south east}{M-13-6.south west}

\DrawCell{M-14-1.north west}{M-14-1.north east}{M-14-1.south east}{M-14-1.south west}
\DrawCell{M-14-2.north west}{M-14-2.north east}{M-14-2.south east}{M-14-2.south west}
\DrawCell{M-14-3.north west}{M-14-3.north east}{M-14-3.south east}{M-14-3.south west}
\DrawCell{M-14-4.north west}{M-14-4.north east}{M-14-4.south east}{M-14-4.south west}
\DrawCell{M-14-5.north west}{M-14-5.north east}{M-14-5.south east}{M-14-5.south west}
\DrawCell{M-14-6.north west}{M-14-6.north east}{M-14-6.south east}{M-14-6.south west}

\DrawCell{M-15-1.north west}{M-15-1.north east}{M-15-1.south east}{M-15-1.south west}
\DrawCell{M-15-2.north west}{M-15-2.north east}{M-15-2.south east}{M-15-2.south west}
\DrawCell{M-15-3.north west}{M-15-3.north east}{M-15-3.south east}{M-15-3.south west}
\DrawCell{M-15-4.north west}{M-15-4.north east}{M-15-4.south east}{M-15-4.south west}
\DrawCell{M-15-5.north west}{M-15-5.north east}{M-15-5.south east}{M-15-5.south west}
\DrawCell{M-15-6.north west}{M-15-6.north east}{M-15-6.south east}{M-15-6.south west}

\node[anchor = north west, align=center, font=\tiny] at (M-2-2.north west) {\\ \textbf{21\%}};
\node[anchor = north west, align=center, font=\tiny] at (M-2-3.north west) {\\ \textbf{57\%}};
\node[anchor = north west, align=center, font=\tiny] at (M-2-4.north west) {\\ \textbf{14\%}};
\node[anchor = north west, align=center, font=\tiny] at (M-2-5.north west) {\\ \textbf{ 7\%}};
\node[anchor = north west, align=center, font=\tiny] at (M-2-6.north west) {\\ \textbf{ 0\%}};

\node[anchor = north west, align=center, font=\tiny] at (M-3-2.north west) {\\ \textbf{ 0\%}};
\node[anchor = north west, align=center, font=\tiny] at (M-3-3.north west) {\\ \textbf{ 0\%}};
\node[anchor = north west, align=center, font=\tiny] at (M-3-4.north west) {\\ \textbf{ 0\%}};
\node[anchor = north west, align=center, font=\tiny] at (M-3-5.north west) {\\ \textbf{35\%}};
\node[anchor = north west, align=center, font=\tiny] at (M-3-6.north west) {\\ \textbf{64\%}};

\node[anchor = north west, align=center, font=\tiny] at (M-4-2.north west) {\\ \textbf{ 0\%}};
\node[anchor = north west, align=center, font=\tiny] at (M-4-3.north west) {\\ \textbf{ 0\%}};
\node[anchor = north west, align=center, font=\tiny] at (M-4-4.north west) {\\ \textbf{14\%}};
\node[anchor = north west, align=center, font=\tiny] at (M-4-5.north west) {\\ \textbf{ 7\%}};
\node[anchor = north west, align=center, font=\tiny] at (M-4-6.north west) {\\ \textbf{78\%}};

\node[anchor = north west, align=center, font=\tiny] at (M-5-2.north west) {\\ \textbf{14\%}};
\node[anchor = north west, align=center, font=\tiny] at (M-5-3.north west) {\\ \textbf{14\%}};
\node[anchor = north west, align=center, font=\tiny] at (M-5-4.north west) {\\ \textbf{21\%}};
\node[anchor = north west, align=center, font=\tiny] at (M-5-5.north west) {\\ \textbf{28\%}};
\node[anchor = north west, align=center, font=\tiny] at (M-5-6.north west) {\\ \textbf{21\%}};

\node[anchor = north west, align=center, font=\tiny] at (M-6-2.north west) {\\ \textbf{25\%}};
\node[anchor = north west, align=center, font=\tiny] at (M-6-3.north west) {\\ \textbf{39\%}};
\node[anchor = north west, align=center, font=\tiny] at (M-6-4.north west) {\\ \textbf{17\%}};
\node[anchor = north west, align=center, font=\tiny] at (M-6-5.north west) {\\ \textbf{17\%}};
\node[anchor = north west, align=center, font=\tiny] at (M-6-6.north west) {\\ \textbf{ 0\%}};

\node[anchor = north west, align=center, font=\tiny] at (M-7-2.north west) {\\ \textbf{50\%}};
\node[anchor = north west, align=center, font=\tiny] at (M-7-3.north west) {\\ \textbf{35\%}};
\node[anchor = north west, align=center, font=\tiny] at (M-7-4.north west) {\\ \textbf{10\%}};
\node[anchor = north west, align=center, font=\tiny] at (M-7-5.north west) {\\ \textbf{ 3\%}};
\node[anchor = north west, align=center, font=\tiny] at (M-7-6.north west) {\\ \textbf{ 0\%}};

\node[anchor = north west, align=center, font=\tiny] at (M-8-2.north west) {\\ \textbf{ 0\%}};
\node[anchor = north west, align=center, font=\tiny] at (M-8-3.north west) {\\ \textbf{ 3\%}};
\node[anchor = north west, align=center, font=\tiny] at (M-8-4.north west) {\\ \textbf{25\%}};
\node[anchor = north west, align=center, font=\tiny] at (M-8-5.north west) {\\ \textbf{32\%}};
\node[anchor = north west, align=center, font=\tiny] at (M-8-6.north west) {\\ \textbf{39\%}};

\node[anchor = north west, align=center, font=\tiny] at (M-9-2.north west) {\\ \textbf{ 0\%}};
\node[anchor = north west, align=center, font=\tiny] at (M-9-3.north west) {\\ \textbf{ 3\%}};
\node[anchor = north west, align=center, font=\tiny] at (M-9-4.north west) {\\ \textbf{ 7\%}};
\node[anchor = north west, align=center, font=\tiny] at (M-9-5.north west) {\\ \textbf{50\%}};
\node[anchor = north west, align=center, font=\tiny] at (M-9-6.north west) {\\ \textbf{39\%}};

\node[anchor = north west, align=center, font=\tiny] at (M-10-2.north west) {\\ \textbf{ 0\%}};
\node[anchor = north west, align=center, font=\tiny] at (M-10-3.north west) {\\ \textbf{ 0\%}};
\node[anchor = north west, align=center, font=\tiny] at (M-10-4.north west) {\\ \textbf{ 0\%}};
\node[anchor = north west, align=center, font=\tiny] at (M-10-5.north west) {\\ \textbf{ 7\%}};
\node[anchor = north west, align=center, font=\tiny] at (M-10-6.north west) {\\ \textbf{92\%}};

\node[anchor = north west, align=center, font=\tiny] at (M-11-2.north west) {\\ \textbf{ 0\%}};
\node[anchor = north west, align=center, font=\tiny] at (M-11-3.north west) {\\ \textbf{ 0\%}};
\node[anchor = north west, align=center, font=\tiny] at (M-11-4.north west) {\\ \textbf{ 3\%}};
\node[anchor = north west, align=center, font=\tiny] at (M-11-5.north west) {\\ \textbf{ 3\%}};
\node[anchor = north west, align=center, font=\tiny] at (M-11-6.north west) {\\ \textbf{92\%}};

\node[anchor = north west, align=center, font=\tiny] at (M-12-2.north west) {\\ \textbf{ 0\%}};
\node[anchor = north west, align=center, font=\tiny] at (M-12-3.north west) {\\ \textbf{ 0\%}};
\node[anchor = north west, align=center, font=\tiny] at (M-12-4.north west) {\\ \textbf{14\%}};
\node[anchor = north west, align=center, font=\tiny] at (M-12-5.north west) {\\ \textbf{39\%}};
\node[anchor = north west, align=center, font=\tiny] at (M-12-6.north west) {\\ \textbf{46\%}};

\node[anchor = north west, align=center, font=\tiny] at (M-13-2.north west) {\\ \textbf{ 0\%}};
\node[anchor = north west, align=center, font=\tiny] at (M-13-3.north west) {\\ \textbf{ 7\%}};
\node[anchor = north west, align=center, font=\tiny] at (M-13-4.north west) {\\ \textbf{21\%}};
\node[anchor = north west, align=center, font=\tiny] at (M-13-5.north west) {\\ \textbf{39\%}};
\node[anchor = north west, align=center, font=\tiny] at (M-13-6.north west) {\\ \textbf{32\%}};

\node[anchor = north west, align=center, font=\tiny] at (M-14-2.north west) {\\ \textbf{10\%}};
\node[anchor = north west, align=center, font=\tiny] at (M-14-3.north west) {\\ \textbf{25\%}};
\node[anchor = north west, align=center, font=\tiny] at (M-14-4.north west) {\\ \textbf{32\%}};
\node[anchor = north west, align=center, font=\tiny] at (M-14-5.north west) {\\ \textbf{28\%}};
\node[anchor = north west, align=center, font=\tiny] at (M-14-6.north west) {\\ \textbf{ 3\%}};

\node[anchor = north west, align=center, font=\tiny] at (M-15-2.north west) {\\ \textbf{32\%}};
\node[anchor = north west, align=center, font=\tiny] at (M-15-3.north west) {\\ \textbf{39\%}};
\node[anchor = north west, align=center, font=\tiny] at (M-15-4.north west) {\\ \textbf{21\%}};
\node[anchor = north west, align=center, font=\tiny] at (M-15-5.north west) {\\ \textbf{ 3\%}};
\node[anchor = north west, align=center, font=\tiny] at (M-15-6.north west) {\\ \textbf{ 3\%}};

\draw[opacity=0.6,fill=OddColor] ($ (M-2-2.north west)!1 - 21 / 100!(M-2-2.south west) $) rectangle(M-2-2.south east);
\draw[opacity=0.6,fill=OddColor] ($ (M-2-3.north west)!1 - 57 / 100!(M-2-3.south west) $) rectangle(M-2-3.south east);
\draw[opacity=0.6,fill=OddColor] ($ (M-2-4.north west)!1 - 14 / 100!(M-2-4.south west) $) rectangle(M-2-4.south east);
\draw[opacity=0.6,fill=OddColor] ($ (M-2-5.north west)!1 - 7 / 100!(M-2-5.south west) $) rectangle(M-2-5.south east);
\draw[opacity=0.6,fill=OddColor] ($ (M-2-6.north west)!1 - 0 / 100!(M-2-6.south west) $) rectangle(M-2-6.south east);

\draw[opacity=0.6,fill=EvenColor] ($ (M-3-2.north west)!1 - 0 / 100!(M-3-2.south west) $) rectangle(M-3-2.south east);
\draw[opacity=0.6,fill=EvenColor] ($ (M-3-3.north west)!1 - 0 / 100!(M-3-3.south west) $) rectangle(M-3-3.south east);
\draw[opacity=0.6,fill=EvenColor] ($ (M-3-4.north west)!1 - 0 / 100!(M-3-4.south west) $) rectangle(M-3-4.south east);
\draw[opacity=0.6,fill=EvenColor] ($ (M-3-5.north west)!1 - 35 / 100!(M-3-5.south west) $) rectangle(M-3-5.south east);
\draw[opacity=0.6,fill=EvenColor] ($ (M-3-6.north west)!1 - 64 / 100!(M-3-6.south west) $) rectangle(M-3-6.south east);

\draw[opacity=0.6,fill=OddColor] ($ (M-4-2.north west)!1 - 0 / 100!(M-4-2.south west) $) rectangle(M-4-2.south east);
\draw[opacity=0.6,fill=OddColor] ($ (M-4-3.north west)!1 - 0 / 100!(M-4-3.south west) $) rectangle(M-4-3.south east);
\draw[opacity=0.6,fill=OddColor] ($ (M-4-4.north west)!1 - 14 / 100!(M-4-4.south west) $) rectangle(M-4-4.south east);
\draw[opacity=0.6,fill=OddColor] ($ (M-4-5.north west)!1 - 7 / 100!(M-4-5.south west) $) rectangle(M-4-5.south east);
\draw[opacity=0.6,fill=OddColor] ($ (M-4-6.north west)!1 - 78 / 100!(M-4-6.south west) $) rectangle(M-4-6.south east);

\draw[opacity=0.6,fill=EvenColor] ($ (M-5-2.north west)!1 - 14 / 100!(M-5-2.south west) $) rectangle(M-5-2.south east);
\draw[opacity=0.6,fill=EvenColor] ($ (M-5-3.north west)!1 - 14 / 100!(M-5-3.south west) $) rectangle(M-5-3.south east);
\draw[opacity=0.6,fill=EvenColor] ($ (M-5-4.north west)!1 - 21 / 100!(M-5-4.south west) $) rectangle(M-5-4.south east);
\draw[opacity=0.6,fill=EvenColor] ($ (M-5-5.north west)!1 - 28 / 100!(M-5-5.south west) $) rectangle(M-5-5.south east);
\draw[opacity=0.6,fill=EvenColor] ($ (M-5-6.north west)!1 - 21 / 100!(M-5-6.south west) $) rectangle(M-5-6.south east);

\draw[opacity=0.6,fill=OddColor] ($ (M-6-2.north west)!1 - 25 / 100!(M-6-2.south west) $) rectangle(M-6-2.south east);
\draw[opacity=0.6,fill=OddColor] ($ (M-6-3.north west)!1 - 39 / 100!(M-6-3.south west) $) rectangle(M-6-3.south east);
\draw[opacity=0.6,fill=OddColor] ($ (M-6-4.north west)!1 - 17 / 100!(M-6-4.south west) $) rectangle(M-6-4.south east);
\draw[opacity=0.6,fill=OddColor] ($ (M-6-5.north west)!1 - 17 / 100!(M-6-5.south west) $) rectangle(M-6-5.south east);
\draw[opacity=0.6,fill=OddColor] ($ (M-6-6.north west)!1 - 0 / 100!(M-6-6.south west) $) rectangle(M-6-6.south east);

\draw[opacity=0.6,fill=EvenColor] ($ (M-7-2.north west)!1 - 50 / 100!(M-7-2.south west) $) rectangle(M-7-2.south east);
\draw[opacity=0.6,fill=EvenColor] ($ (M-7-3.north west)!1 - 35 / 100!(M-7-3.south west) $) rectangle(M-7-3.south east);
\draw[opacity=0.6,fill=EvenColor] ($ (M-7-4.north west)!1 - 10 / 100!(M-7-4.south west) $) rectangle(M-7-4.south east);
\draw[opacity=0.6,fill=EvenColor] ($ (M-7-5.north west)!1 - 3 / 100!(M-7-5.south west) $) rectangle(M-7-5.south east);
\draw[opacity=0.6,fill=EvenColor] ($ (M-7-6.north west)!1 - 0 / 100!(M-7-6.south west) $) rectangle(M-7-6.south east);

\draw[opacity=0.6,fill=OddColor] ($ (M-8-2.north west)!1 - 0 / 100!(M-8-2.south west) $) rectangle(M-8-2.south east);
\draw[opacity=0.6,fill=OddColor] ($ (M-8-3.north west)!1 - 3 / 100!(M-8-3.south west) $) rectangle(M-8-3.south east);
\draw[opacity=0.6,fill=OddColor] ($ (M-8-4.north west)!1 - 25 / 100!(M-8-4.south west) $) rectangle(M-8-4.south east);
\draw[opacity=0.6,fill=OddColor] ($ (M-8-5.north west)!1 - 32 / 100!(M-8-5.south west) $) rectangle(M-8-5.south east);
\draw[opacity=0.6,fill=OddColor] ($ (M-8-6.north west)!1 - 39 / 100!(M-8-6.south west) $) rectangle(M-8-6.south east);

\draw[opacity=0.6,fill=EvenColor] ($ (M-9-2.north west)!1 - 0 / 100!(M-9-2.south west) $) rectangle(M-9-2.south east);
\draw[opacity=0.6,fill=EvenColor] ($ (M-9-3.north west)!1 - 3 / 100!(M-9-3.south west) $) rectangle(M-9-3.south east);
\draw[opacity=0.6,fill=EvenColor] ($ (M-9-4.north west)!1 - 7 / 100!(M-9-4.south west) $) rectangle(M-9-4.south east);
\draw[opacity=0.6,fill=EvenColor] ($ (M-9-5.north west)!1 - 50 / 100!(M-9-5.south west) $) rectangle(M-9-5.south east);
\draw[opacity=0.6,fill=EvenColor] ($ (M-9-6.north west)!1 - 39 / 100!(M-9-6.south west) $) rectangle(M-9-6.south east);

\draw[opacity=0.6,fill=OddColor] ($ (M-10-2.north west)!1 - 0 / 100!(M-10-2.south west) $) rectangle(M-10-2.south east);
\draw[opacity=0.6,fill=OddColor] ($ (M-10-3.north west)!1 - 0 / 100!(M-10-3.south west) $) rectangle(M-10-3.south east);
\draw[opacity=0.6,fill=OddColor] ($ (M-10-4.north west)!1 - 0 / 100!(M-10-4.south west) $) rectangle(M-10-4.south east);
\draw[opacity=0.6,fill=OddColor] ($ (M-10-5.north west)!1 - 7 / 100!(M-10-5.south west) $) rectangle(M-10-5.south east);
\draw[opacity=0.6,fill=OddColor] ($ (M-10-6.north west)!1 - 92 / 100!(M-10-6.south west) $) rectangle(M-10-6.south east);

\draw[opacity=0.6,fill=EvenColor] ($ (M-11-2.north west)!1 - 0 / 100!(M-11-2.south west) $) rectangle(M-11-2.south east);
\draw[opacity=0.6,fill=EvenColor] ($ (M-11-3.north west)!1 - 0 / 100!(M-11-3.south west) $) rectangle(M-11-3.south east);
\draw[opacity=0.6,fill=EvenColor] ($ (M-11-4.north west)!1 - 3 / 100!(M-11-4.south west) $) rectangle(M-11-4.south east);
\draw[opacity=0.6,fill=EvenColor] ($ (M-11-5.north west)!1 - 3 / 100!(M-11-5.south west) $) rectangle(M-11-5.south east);
\draw[opacity=0.6,fill=EvenColor] ($ (M-11-6.north west)!1 - 92 / 100!(M-11-6.south west) $) rectangle(M-11-6.south east);

\draw[opacity=0.6,fill=OddColor] ($ (M-12-2.north west)!1 - 0 / 100!(M-12-2.south west) $) rectangle(M-12-2.south east);
\draw[opacity=0.6,fill=OddColor] ($ (M-12-3.north west)!1 - 0 / 100!(M-12-3.south west) $) rectangle(M-12-3.south east);
\draw[opacity=0.6,fill=OddColor] ($ (M-12-4.north west)!1 - 14 / 100!(M-12-4.south west) $) rectangle(M-12-4.south east);
\draw[opacity=0.6,fill=OddColor] ($ (M-12-5.north west)!1 - 39 / 100!(M-12-5.south west) $) rectangle(M-12-5.south east);
\draw[opacity=0.6,fill=OddColor] ($ (M-12-6.north west)!1 - 46 / 100!(M-12-6.south west) $) rectangle(M-12-6.south east);

\draw[opacity=0.6,fill=EvenColor] ($ (M-13-2.north west)!1 - 0 / 100!(M-13-2.south west) $) rectangle(M-13-2.south east);
\draw[opacity=0.6,fill=EvenColor] ($ (M-13-3.north west)!1 - 7 / 100!(M-13-3.south west) $) rectangle(M-13-3.south east);
\draw[opacity=0.6,fill=EvenColor] ($ (M-13-4.north west)!1 - 21 / 100!(M-13-4.south west) $) rectangle(M-13-4.south east);
\draw[opacity=0.6,fill=EvenColor] ($ (M-13-5.north west)!1 - 39 / 100!(M-13-5.south west) $) rectangle(M-13-5.south east);
\draw[opacity=0.6,fill=EvenColor] ($ (M-13-6.north west)!1 - 32 / 100!(M-13-6.south west) $) rectangle(M-13-6.south east);

\draw[opacity=0.6,fill=OddColor] ($ (M-14-2.north west)!1 - 10 / 100!(M-14-2.south west) $) rectangle(M-14-2.south east);
\draw[opacity=0.6,fill=OddColor] ($ (M-14-3.north west)!1 - 25 / 100!(M-14-3.south west) $) rectangle(M-14-3.south east);
\draw[opacity=0.6,fill=OddColor] ($ (M-14-4.north west)!1 - 32 / 100!(M-14-4.south west) $) rectangle(M-14-4.south east);
\draw[opacity=0.6,fill=OddColor] ($ (M-14-5.north west)!1 - 28 / 100!(M-14-5.south west) $) rectangle(M-14-5.south east);
\draw[opacity=0.6,fill=OddColor] ($ (M-14-6.north west)!1 - 3 / 100!(M-14-6.south west) $) rectangle(M-14-6.south east);

\draw[opacity=0.6,fill=EvenColor] ($ (M-15-2.north west)!1 - 32 / 100!(M-15-2.south west) $) rectangle(M-15-2.south east);
\draw[opacity=0.6,fill=EvenColor] ($ (M-15-3.north west)!1 - 39 / 100!(M-15-3.south west) $) rectangle(M-15-3.south east);
\draw[opacity=0.6,fill=EvenColor] ($ (M-15-4.north west)!1 - 21 / 100!(M-15-4.south west) $) rectangle(M-15-4.south east);
\draw[opacity=0.6,fill=EvenColor] ($ (M-15-5.north west)!1 - 3 / 100!(M-15-5.south west) $) rectangle(M-15-5.south east);
\draw[opacity=0.6,fill=EvenColor] ($ (M-15-6.north west)!1 - 3 / 100!(M-15-6.south west) $) rectangle(M-15-6.south east);
\end{tikzpicture}
\footnotesize A response of 1 corresponds to strongly disagree whilst a response of 5 corresponds to strongly agree.
\end{figure}

When asked about their most preferred aspect of the assistant 29\% of participants mentioned the subtleness of assistance provided stating that it was “non-invasive and well integrated with manual control” whilst some participants “did not notice it was there”. A further 29\% of participants were most appreciative with the lack of effort required to land by having to only approximately approach the platform while 21\% of participants enjoyed the improved success rate. 

\begin{table}
\begin{center}
\captionof{table}{Final survey worded questions}
\label{WordedResponse}
\begin{tabular}{|m{7.2cm}|}
  \hline
  \textbf{1:} What part of the task did you have the most difficulties with?\\
  \hline
  \textbf{2:} What additional features or changes would make the task easier?\\
  \hline
  \textbf{3:} What do you think the assistant was doing during the study?\\
  \hline
  \textbf{4:} What did you like the most about the assistant?\\
  \hline
  \textbf{5:} What did you dislike the most about the assistant?\\
  \hline
  \textbf{6:} Describe differences in your flying strategy when unassisted/assisted\\
  \hline
  \textbf{7:} Any additional comments or observations you would like to add?\\
  \hline
\end{tabular}
\end{center}
\end{table}

When asked about their most disliked aspect of the assistant 21\% of participants made references to the uncertainty in the intent of the assistant stating that “I wasn’t sure what it was doing” and that “it was completely invisible”. Although many participants’ most enjoyed aspect of the assistant was its subtleness, this became a double-edged sword for others where the lack of feedback and communication resulted in a sense of uncertainty in what was expected of themselves and the assistant. 18\% of participants stated that they thought the assistant landed the drone with too great of a 

vertical velocity and a further 18\% of participants made comments about disliking the assistant’s automatic landing when close to the platform. Those participants made comments of wanting to be able to “readjust the position of the drone” when close to the platform and preferred if the assistant “only controlled the x-y position of the drone”, giving full control of the throttle to the participant.

\begin{table}[b]
\footnotesize
\begin{center}
\captionof{table}{Participant proficiency regression results}
\label{ProficiencyRegression}
\setlength\tabcolsep{2.5pt}
\begin{tabular}{ |l|c|c|c|c|c|c| } 
\hline
\multicolumn{1}{|l|}{Independent variables} & \multicolumn{5}{c|}{P-values} & \multicolumn{1}{c|}{Coef. (m)}\\
\hline
 \rowcolor{TableGray} Intercept & 0.00 & 0.00 & 0.00 & 0.00 & 0.00 & 0.378 \\ 
 Drone exp. & 0.00 & 0.00 & 0.00 & 0.00 & 0.00 & -0.030 \\ 
 \rowcolor{TableGray} Joystick exp. & 0.01 & 0.01 & 0.01 & 0.00 & 0.00 & -0.025 \\ 
 Video game frequency & 0.06 & 0.03 & 0.03 & 0.03 & 0.03 & 0.016 \\ 
 \hline
 \rowcolor{TableGray} Drone confidence & 0.33 & 0.33 & 0.25 & 0.27 & \textemdash & \textemdash \\ 
 Flying game exp. & 0.61 & 0.58 & 0.57 & \textemdash & \textemdash & \textemdash \\ 
 \rowcolor{TableGray} Flown before & 0.81 & 0.83 & \textemdash & \textemdash & \textemdash & \textemdash \\ 
 Unassisted second & 0.87 & \textemdash & \textemdash & \textemdash & \textemdash & \textemdash \\ 
 \hline
\end{tabular}
\end{center}
\end{table}

To measure the impact of participants' prior experience and to determine whether the assistant allows novice pilots to perform the task at a proficiency equal to or greater than experienced pilots, a multi-variable linear regression model was established to assess  participants' expertise level. Participants’ responses from the demographic survey as seen in Table.~\ref{DemographicsTable} were used as independent variables in the model including an additional independent variable to denote whether the participant performed the unassisted task in the second half to control for potential learning throughout the study. The regression model used the participants’ average unassisted landing error as the dependent variable where multiple regressions were performed using backwards elimination until all remaining independent variables were deemed statistically significant under a two-tailed t-test with significance level of \(\alpha\) = 0.05. The results of the backwards elimination regression can be seen in Table.~\ref{ProficiencyRegression}.

Prior drone piloting experience, joystick experience and video game frequency independent variables were found to be statistically significant under \(\alpha = 0.05\). The model suggests that participants with a greater piloting and joystick experience on average had a lower unassisted landing error than those that did not, while those who frequently played video games tended to perform worse, which may possibly be caused from unrealistic expectations of real-world UAV dynamics. Insufficient evidence exists to suggest that the remaining independent variables had an effect on participants’ performance.  The order in which the participants performed the task had no significant effect on their unassisted landing results. 

\begin{figure}[b]
\centering
\includegraphics[width=1.0\columnwidth]{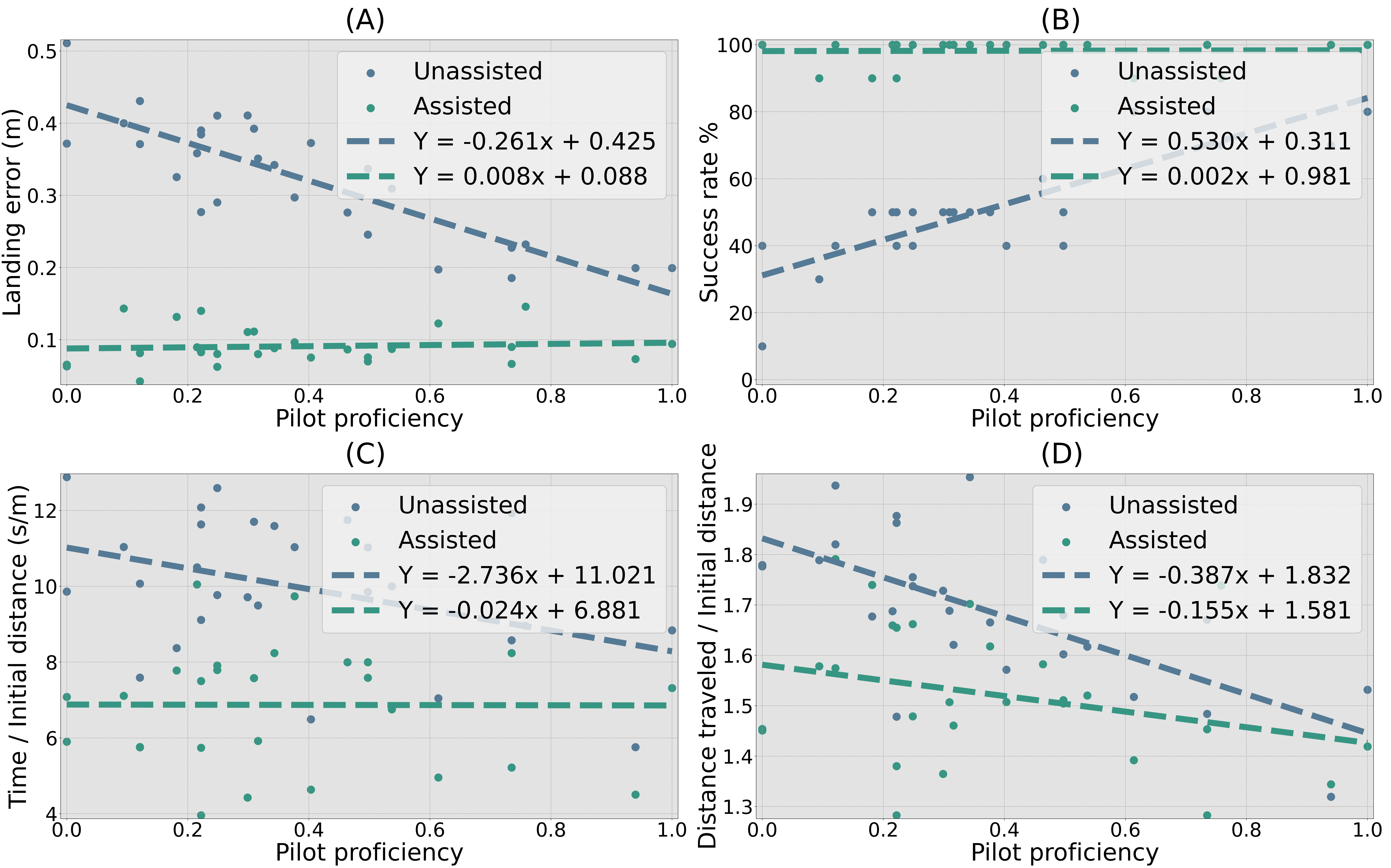}
\caption{Plots of participants’ averaged performance metrics against proficiency for unassisted and assisted conditions.}
\label{RegressionImage}
\end{figure}

To determine a participant's proficiency score, the regressed model was used to estimate their unassisted landing error which was then remapped to fit within the range \([0, 1]\), where a value of zero denotes a novice pilot whilst a value of one denotes an expert pilot. The resultant proficiency score can be plotted against key performance metrics to measure whether pilot proficiency has an effect under the unassisted and assisted conditions, the results of which can be seen in Fig. \ref{RegressionImage}.

Plot A in Fig. \ref{RegressionImage} shows a strong negative correlation between participants’ proficiency and their unassisted landing error, which intuitively leads to higher unassisted success rates in plot B. This strong relationship is expected as the previous regression model for estimating participants’ proficiency regressed against their unassisted landing error. For the assisted condition it can be seen that participants’ landing error and success rate was invariant to their prior piloting experience. Regardless of proficiency, participants performed consistently in the assisted condition, at a level greater than that of the highest performing participants. 

For efficiency metrics in plots C and D in Fig. \ref{RegressionImage}, a negative correlation exists between participants’ proficiency and both the time taken to complete the task and the distance traveled. Meaning that on average the more proficient a participant is, the shorter the flight trajectory and time that is required to complete the task. Whilst in the assisted condition, the average time taken did not depend on the participants’ proficiency, but a weak negative correlation is observed for the trajectory distance.

To confirm whether a statistical relationship exists between the aforementioned metrics and participant proficiency, the slope coefficients are subjected to a two-tailed t-test under significance \(\alpha = 0.05\). For the unassisted condition, sufficient evidence exists to suggest that participant proficiency does affect all metrics shown in Fig. \ref{RegressionImage}. Whilst for the assisted condition, there is insufficient evidence to suggest that participant proficiency influences any of the tested metrics and that regardless of ones prior piloting experience, participants are projected to perform equally across all metrics with the help of the assistant. 

\begin{figure*}[ht]
\centering
\includegraphics[width=2.0\columnwidth]{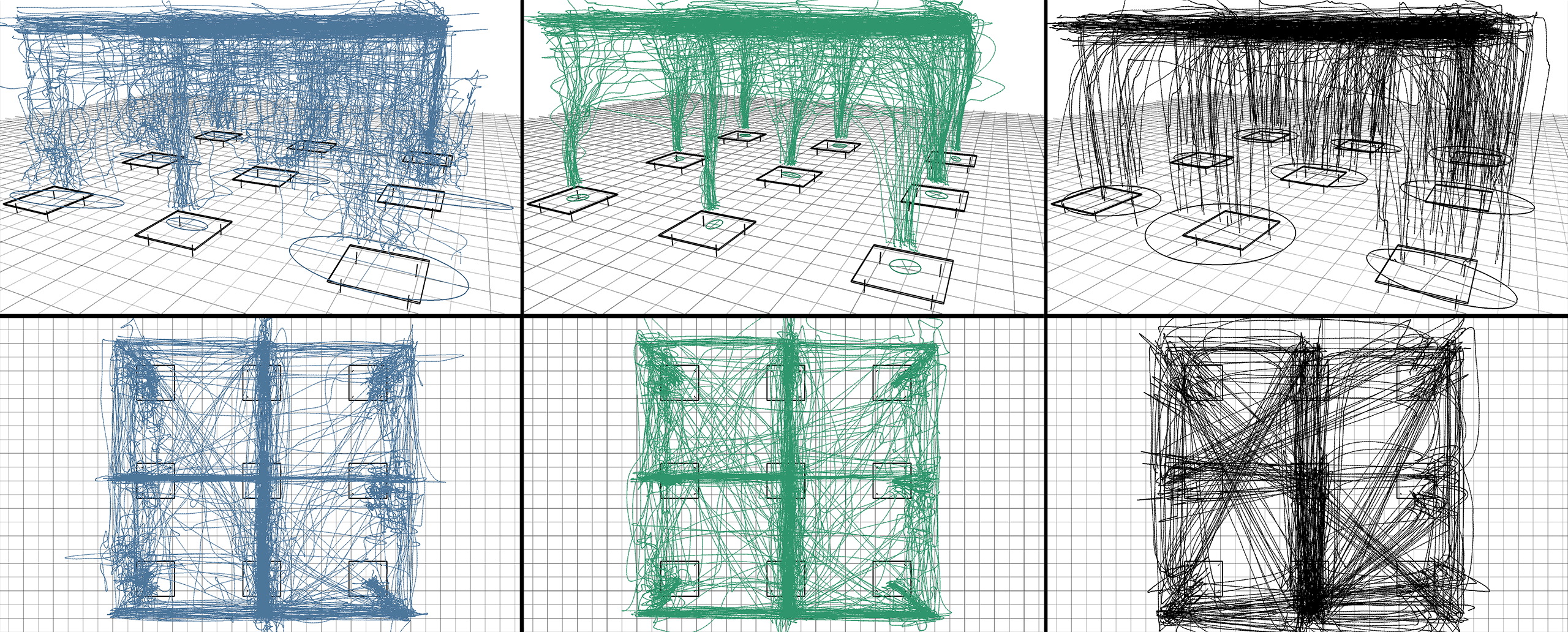}
\caption{Total trajectories from user study in the unassisted condition (left) and assisted condition (middle), alongside an equivalent sample of randomly selected simulated users (right) performing the same task in the simulated validation environment. Ellipses are centered at the average landing location where principal component analysis is performed to demonstrate the axis with greatest deviation for each landing platform.}
\label{TotalTrajectories}
\end{figure*}

During the assisted portion of the user study, it was observed that participants developed riskier or lower-effort flying habits. In the later assisted landings participants provided low-effort initial estimates of the target platform compared to earlier assisted landings. Participants also made comments about becoming “reckless” in their post-study responses, which was observed by participants flying at greater speeds with aggressive descents in the later portion of assisted landings. In general, as participants’ confidence in the assistant grew, the level of risk they were willing to take also grew. The observations of low-effort initial estimates and increased risk-taking behavior are empirically supported through the final four landings in the assisted condition. A statistically significant difference in the landing error between the initial (0.08m) and later portion (0.11m) of the assisted landings were observed using a Welch’s t-test (p=0.008) under a 95\% significance level. Of the five failed assisted landings, 80\% occurred in the final four landings. Performance degradation was not observed in the unassisted condition where participants performed better in the later portion of the unassisted landings.

\section{Discussion}

The success of our proposed shared autonomy approach can be attributable to three main factors: (i) domain randomization, (ii) simulated user modeling and (iii) training efficiency. Transferring models trained purely in simulation must pass the simulation-to-reality gap, where visual perception accounts for the most significant portion of the gap due to difficulties in replicating photorealistic images, leading to simulation trained models to fail once transferred \citep{RoboticGrasping}. Domain randomization reduces the need for photorealistic images with the aim of making the trained network invariant to unimportant information i.e., illumination and background textures, which has shown success in other UAV applications \citep{DeepDroneRacing}. By training the perception module over a wide variety of simulated scenes under intense dynamically generated noise, the perception module becomes robust to visual disturbances. This is reflected in the results in Section \ref{PerceptionResults}, where the perception module trained with high domain randomization was found to have lower reconstruction errors when transferred to real scenes. Domain randomization is also a key consideration when developing simulated users as human pilots vary greatly in terms of strategy and proficiency.

Developing simulated users in a shared autonomy system is the most difficult challenge as they must cover a broad range of policies to reflect variability amongst users and realistically react to actions taken by the actor. Focusing the development of the simulated user around four base parameters simplifies the task while being capable of displaying a wide range of policies through the variation of the base parameters. Fig. \ref{TotalTrajectories} shows the experimentally observed flight trajectories of participants in the unassisted and assisted conditions alongside an equivalent sample of unassisted simulated users. Participants approached the target platform in two ways, by flying along the principle axes or by flying directly to the platform as modeled by the simulated user. During descent unassisted participants appear to follow a complex trajectory, which is unmodeled by the simulated user. During the assisted condition, the trajectories reflect the smoothness of the simulated user’s model but with lower landing error variance. The source of trajectory complexity in the unassisted condition stems from uncertainty felt by the participants, which is absent in the assisted condition leading to simpler trajectories. The robustness of the user model is also illustrated in Fig. \ref{ComparisonImage}, which shows that the proposed model achieves the best performance even when tested against unseen user models.

The third factor which made our shared autonomy approach successful was training efficiency. From the ablation studyresults in Section \ref{Result_TD3}, it was found that providing additional information only available in simulation to the critic greatly improved convergence time. Training the actor took approximately 1.5 days to complete with 16 UAVs flying concurrently. Our previous work \citep{Kal} did not include additional information to the critic or perform concurrent exploration. Instead, it heavily relied on preloading the experience replay buffer with prior failed models’ state-transitions and performing additional optimization iterations to converge within a reasonable time. The training efficiency made it practical to perform hyper-parameter optimization, reward structure modification and ensured proper model convergence before being deployed. 

A major focus of the proposed approach was in developing an assistant that was subtle and non-intrusive by implicitly inferring the pilot’s intent and only providing assistance as needed. From the responses outlined in the user study results in Section \ref{UserStudyResults}, it was found that some participants preferred this non-intrusive behavior for its intuitive controls, however others found the lack of communication and feedback unsettling. This may be due to participants not being told how the assistant operates or because of only using the assistant for a short period of time, however further work needs to be done investigating if a non-intrusive approach or one that actively communicates to the pilot is most appropriate. 

Another concern is the observed behavioral change participants displayed in the later portion of the assisted condition with decreased effort and increased risk-taking, expecting the assistant to cover for them. Although the role of the assistant is to ensure safe landing with decreased effort and requisite skill, prolonged exposure may lead to degraded piloting proficiency over extended use. However the decrease in performance may be due to participants being tested in a low-stakes environment where the curiosity of exploring the limits and behavior of the assistant became more interesting.

A limitation of the proposed work is the reliance on motion capture for pose estimation when flying in the GPS denied environment. To extend the proposed approach to outdoor conditions, additional onboard sensing would need to be implemented to fuse GPS, IMU and visual odometry sensor streams. Visual odometry information can be attainable using the Intel RealSense Tracking camera T265 which has shown success in autonomous UAV landing tasks \citep{Realsense1, Realsense2}.    

Although the proposed approach has been fine tuned for UAV landing tasks, the principles of the approach have the potential to be transferred to alternative tasks. Three components are required for replication in alternative environments: (1) a perception module that perceives the environment by encoding information onto a latent vector, achievable with any auto-encoder network architecture trained to reconstruct or segment images. (2) A parametric simulated user model, where the parameters account for adaptability towards the assistant’s actions (\(\alpha\)), the proficiency of the simulated user in its ability to successfully complete the tasks by itself (\(\beta\)) and parameters associated with the dynamics of the output control variables (\(\psi\) \&  \(\phi\)). The third component required is a policy module trained using TD3 with critic MDP formulation that includes the simulated user's objective, while reward terms focus on task success components (\(R_\mathrm{LandingError}\), \(R_\mathrm{SafePos}\), \(R_\mathrm{HVel}\) \& \(R_\mathrm{VVel}\)) and agreement with the user (\(R_\mathrm{ActionDiff}\)).

\section{Conclusion}
In this work we propose a shared autonomy approach that assists pilots of all skill levels to safely land a UAV under conditions where depth perception is difficult. The assistant is unaware of the pilot’s intent nor any prior knowledge of the structure of the environment and must disambiguate the task using observations of the pilot’s actions and its immediate surroundings. The proposed approach comprises of two fundamental components (i) a perception module that encodes information about the structure of the environment from two RGB-D cameras and (ii) a policy network that provides control inputs to assist the pilot in safely landing the UAV.

A user study ($n=28$) was held to validate the assistant’s performance where participants were instructed to land the drone on one of nine platforms in both the unassisted and assisted conditions. With the help of the assistant participants had significantly higher success rates from 51.4\% to 98.2\% and completed the task more efficiently in respect to time taken and distance traveled. Regression analysis showed that regardless of prior experience, the assistant allowed participants of all piloting proficiencies to perform equally amongst each other and greater than that of the most proficient unassisted pilots. 

Compared to the unassisted condition, participants perceived the assisted condition to require a lower task load and to be the more favorable approach according to the NASA TLX and final survey. Participants stated that their most preferred aspect of the assistant was its non-invasive subtle nature as well as an overall reduction in effort required to successfully complete the task. However the assistant’s subtleness was a divisive aspect for others where the lack of communication and inability to give feedback about its intent led to participants feeling a sense of uncertainty in what was expected of them. 
Further work needs to be done to address how an AI in a shared autonomy setup can effectively communicate its intent and provide feedback on the user’s actions whilst maintaining an intuitive non-intrusive behavior. 

\section{Ethical statement}
The authors declare that the submitted work is free from personal conflicts of interest.
The user study was approved by the Monash University Human Research Ethics Committee (MUHREC), project ID 29565. All participants gave informed consent prior to participating in the user study.

\section{Author contribution}
Project Concept: [Kal Backman, Hoam Chung, Dana Kulic]; Methodology: [Kal Backman, Dana Kulic];  User Study Design and Implementation  [Kal Backman, Hoam Chung, Dana Kulic]; User Study Execution and Data Acquisition  [Kal Backman]; Analysis of the Results  [Kal Backman, Hoam Chung, Dana Kulic]; Writing - original draft preparation: [Kal Backman]; Writing - review and editing: [Hoam Chung, Dana Kulic]; Supervision: [Hoam Chung, Dana Kulic]

\section{Acknowledgements}
None

\section{Funding}
This work was supported in part by D. Kuli\'c's Australian Research Council Future Fellowship (FT200100761).

\end{document}